\def\year{2021}\relax
\documentclass[letterpaper]{article} 
\usepackage{aaai21}  
\usepackage{times}  
\usepackage{helvet} 
\usepackage{courier}  
\usepackage[hyphens]{url}  
\usepackage{graphicx} 
\def\UrlFont{\rm}  
\usepackage{natbib}  
\usepackage{caption} 
\usepackage{amsmath}
\usepackage{amsthm}
\usepackage{amssymb}
\usepackage{xcolor}
\usepackage{graphicx}
\usepackage{wrapfig}
\usepackage{xfrac}
\usepackage[]{todonotes}
\usepackage{pifont}
\usepackage{appendix}
\usepackage{euscript}
\usepackage[linesnumbered, ruled,vlined]{algorithm2e}
\usepackage{booktabs}
\usepackage{siunitx}
\usepackage{etoolbox}
\usepackage{multirow}

\newcommand{\tinypm}{\raisebox{1.0pt}{$_{^\pm}$}}
\newcommand{\ubold}{\fontseries{b}\selectfont}
\newcommand{\usecond}{\fontseries{b}\fontshape{it}\selectfont}
\robustify\ubold
\usepackage{upgreek}

\usepackage{adjustbox}

\allowdisplaybreaks

\definecolor{figblue}{HTML}{3E4AFF}
\definecolor{figred}{HTML}{FF2943}
\definecolor{figpurple}{HTML}{DA2FFD}

\title{Message Passing Neural Processes}
\author{
Ben Day*, C\u{a}t\u{a}lina Cangea*, Arian Jamasb, Pietro Li\`{o}\\
}
\affiliations{
Department of Computer Science and Technology, University of Cambridge
}

\begin{document}

\maketitle

\begin{abstract}
Neural Processes (NPs) are powerful and flexible models able to incorporate uncertainty when representing stochastic processes, while maintaining a linear time complexity.
However, NPs produce a latent description by aggregating independent representations of context points and lack the ability to exploit relational information present in many datasets.
This renders NPs ineffective in settings where the stochastic process is primarily governed by neighbourhood rules, such as cellular automata (CA), and limits performance for any task where relational information remains unused.
We address this shortcoming by introducing Message Passing Neural Processes (MPNPs), the first class of NPs that explicitly makes use of relational structure within the model.
Our evaluation shows that MPNPs thrive at lower sampling rates, on existing benchmarks and newly-proposed CA and Cora-Branched tasks.
We further report strong generalisation over density-based CA rule-sets and significant gains in challenging arbitrary-labelling and few-shot learning setups.
\end{abstract}

\section{Introduction}

\noindent Neural Networks (NNs) are a class of methods widely adopted in single-task learning scenarios, where large quantities of labelled data are available. They exhibit favourable properties such as $\mathcal{O}(|D|)$ prediction time complexity, where $D$ is the sample set. However, they are harder to adapt to challenging scenarios, such as multi-task or few-shot learning, and do not typically provide uncertainty estimates for predictions. Relational inductive biases have been added to NNs \cite{battaglia2018relational}, producing models called Graph Neural Networks that are able to exploit relational information via message-passing operations. Alternatively, Gaussian Processes (GPs) \citep{williams1996gaussian} are better suited to non-standard tasks and estimate uncertainty, albeit at often unscalable prediction costs ($\mathcal{O}(|D|^3)$).

Neural Processes (NPs)~\cite{garnelo2018neural} aim to combine the best of both worlds: they learn to represent a stochastic process using labelled samples from its instantiations, with a global latent variable modelling the stochasticity of the learned functions. At test time, only a few labelled points are required to produce predictions for the rest of the dataset, along with their associated uncertainties, in linear time. These models have been successful in few-shot learning and multi-task settings \citep{garnelo2018neural, garnelo2018conditional, requeima2019fast}, but do not leverage the structural information in the data, an approach which has been highly effective on relational tasks \citep{ZhouGraphApplications}. To address this limitation, we propose a novel Neural Process model for classification, which explicitly incorporates structural information when modelling stochastic processes, and showcase the benefits of our method on a wide variety of benchmarks. In this way, our modifications parallel those of the Convolutional Conditional Neural Process (ConvCNP) \cite{gordon2019convolutional}, which also equips NP models with a better inductive bias to build richer representations of the available context.

The central contribution of our work is the \emph{Message Passing Neural Process} (MPNP), the first node classification framework that learns to represent stochastic processes which yield datasets with explicit relational information. We experimentally validate the relative strengths of MPNPs on a variety of existing geometric and biological tasks. In addition, we introduce a challenging new collection of Cellular Automata datasets to test the ability of the model to handle broad variation in the function distribution, where most existing baselines fail to achieve better-than-chance performance. Finally, we construct \emph{Cora-Branched}---a set of novel arbitrary-labelling and few-shot learning tasks based on the Cora dataset---and again show significant MPNP gains.

\section{Background and Related Work}
\label{s:background}

We begin by reviewing the theoretical foundations of our building blocks (Neural Processes, Message Passing architectures) and related works. The next section presents MPNPs as a combination of these ideas that operates on datasets with relational structure generated by stochastic processes.

\subsection{Neural Processes}

\subsubsection{Problem Statement} Given a set of points with features $X$, partially labelled by a function $f:X\rightarrow Y$ sampled from a distribution over functions, $\mathcal{D}$, the goal is to predict labels for a subset of the unlabelled points.

A Neural Process (NP)~\cite{garnelo2018neural} learns to represent a stochastic process with an underlying distribution $\mathcal{D}$. To achieve this, the NP is trained on a set of functions $f : X \rightarrow Y$ sampled from $\mathcal{D}$ and tested on a disjoint set. For each function, $f_i$, a dataset contains tuples $(x_j, y_j)$, where $y_j = f_i(x_j)$. Their joint probability distribution can be written as $p(y_{1:n} | x_{1:n}) = \int p(f_i) p(y_{1:n} | f_i, x_{1:n}) df_i$. Assuming observation noise $Y_j \sim \mathcal{N}(f_i(x_j), \sigma^2)$ and a neural network $\gamma$ modelling the stochastic process instance $f_i$ (that is, $\gamma(x, z) = f_i(x)$, where $z$ is a random vector that mimics the randomness of $f_i$), we obtain the generative model:
\begin{equation}\label{eq:npgen}
    p(z, y_{1:n} | x_{1:n}) = p(z) \prod_{j=1}^n \mathcal{N}(y_j | \gamma(x_j, z), \sigma^2),
\end{equation}
where $p(z)$ is a multivariate normal distribution. Learning the non-linear function $\gamma$ requires amortised variational inference on the evidence lower bound (ELBO), using a neural-network-parameterised posterior $q(z | x_{1:n}, y_{1:n})$. Model generation starts with the NP receiving a set of $m$ context points $\mathcal{C} = \{(x_j, y_j)\}_{j=1}^{m}$ sampled from $f_i$. The model then predicts the values $y_j = f_i(x_j)$ for $n$ target points $\mathcal{T} = \{x_j\}_{j=1}^{n}$; namely, the $m$ original context points and $m-n$ previously unseen target points. To match this setup, we further isolate the context set $x_{1:m}, y_{1:m}$ from the target set $x_{m+1:n}, y_{m+1:n}$ in equation~\ref{eq:npgen}. The final ELBO is:
\begin{multline}
    \log{p(y_{m+1:n} | x_{1:n}, y_{1:m})} \geq \\
    \mathbb{E}_{q(z | x_{1:n}, y_{1:n})} \Big[ \sum_{j=m+1}^n \log{p(y_j | z, x_j)} \\ + \log{\frac{q(z | x_{1:m}, y_{1:m})}{q(z | x_{1:n}, y_{1:n})}} \Big].
\end{multline}
Crucially, NPs are trained on multiple datasets (i.e. sets of samples from functions $f_i$), to provide information about the variability of the stochastic process that is being modelled. 

\subsection{Message Passing and Graph Neural Networks}
Neural networks that operate on graph-structured data process node features $\mathbf{X} \in \mathbb{R}^{n \times d}$ with the relational information in the form of an adjacency matrix $\mathbf{A} \in \{0, 1\}^{n \times n}$. The aim is to produce embeddings that are useful for downstream tasks such as node or graph classification. Graph neural networks typically use generalised convolutional layers to learn these embeddings. We describe their operation via the universal Message Passing (MP) paradigm; the next section presents the specific MP instance that our models use.

Assume $\mathbf{h}_{i}^{t} \in \mathbb{R}^{d'}$ to be the features of the $i$-th node after $t$ message passing steps, where $d'$ is the embedding dimensionality; optionally, we may have edge features $\mathbf{e}_{ij} \in \mathbb{R}^k$ for any $i, j$ where $A_{ij} = 1$. A message passing layer corresponds to a single message passing step, updating the node features as follows, where $F$ and $G$ are learnable functions, $\EuScript{N}(i) = \{j ~|~ A_{ij} = 1\}$ and $\square$ is a permutation-invariant aggregation function:
\begin{equation} \label{eq:mp}
    \mathbf{h}_{i}^{t+1} = \textit{MP}(\mathbf{h}^t) \triangleq F(\mathbf{h}_{i}^t, \square {}_{j \in \EuScript{N}(i)}, G(\mathbf{h}_{i}^t, \mathbf{h}_{j}^t, \mathbf{e}_{ij})).
\end{equation}

\subsubsection{Neural Process Models} \citet{garnelo2018neural} formulated the \emph{Neural Process} as a favourable combination of neural networks and Gaussian Processes. \emph{Conditional Neural Processes} (CNPs)~\cite{garnelo2018conditional} are NP instances without a global latent variable, which implies a deterministic dependence on the context set. \emph{Attentive NPs}~\cite{kim2019attentive}, \emph{CNAPs} \cite{requeima2019fast}, \emph{Convolutional CNPs}~\cite{gordon2019convolutional} and \emph{Sequential NPs}~\cite{NIPS2019_9214} make modifications to reduce underfitting, better adapt in the multi-task setting, and apply inductive biases for translation and temporal sequences, respectively. \citet{NIPS2019_9079} propose the \emph{Functional NP} that learns a graph of dependencies between latent representations of the points, without placing a prior over the latent global variable, though their tasks do not contain explicit relational information. The \emph{Graph NP}~\cite{Carr2019GraphNetworks} is most closely related to the MPNP, performing edge imputation using a CNP-based model and Laplacian-derived features for the context points. However, despite the naming similarity, Graph NPs and MPNPs address different tasks---the former was evaluated on link prediction tasks, which is not in the scope of our work. Moreover, our NP-based model is more flexible, handles uncertainty and learns from neighbourhoods, rather than whole-graph features, for classifying individual dataset samples (nodes), while leveraging the structure between them (edges).

\subsubsection{Graph Learning under Uncertainty} \emph{Graph Gaussian Processes}~\cite{ng2018bayesian} were designed as an extension to GPs, where the covariance function and prior exploit the existence of features in node neighbourhoods. Graph GPs are the only Gaussian method for node classification, but perform slightly worse than GCNs---a type of GNNs that we use as a baseline. Moreover, the complexity is somewhat higher: $\mathcal{O}(\texttt{max\_node\_degree}^2 * N)$ vs. $\mathcal{O}(N)$ for (MP)NP, where $N =$ set of observations/context nodes. The \emph{Relational GP}~\cite{chu2007relational} models pairwise undirected links between data points, thus addressing a different task. The \emph{Graph Convolutional GP}~\cite{walker2019graph} is a translation-invariant model that operates similarly to convolutional layers, while generalising to non-Euclidean domains. More recently,~\citet{opolka2020graph} have also proposed a \emph{Graph Convolutional GP} model for link prediction, which uses a GP for node-level predictions, another GP that builds on the first one for edge-level predictions, and a deep GP incorporating these building blocks to produce more expressive representations.

\section{Message Passing Neural Processes} \label{s:mpnps}
\begin{figure*}[t]
    \centering
    \includegraphics[trim= 10 69 173 10, clip, width=\textwidth]{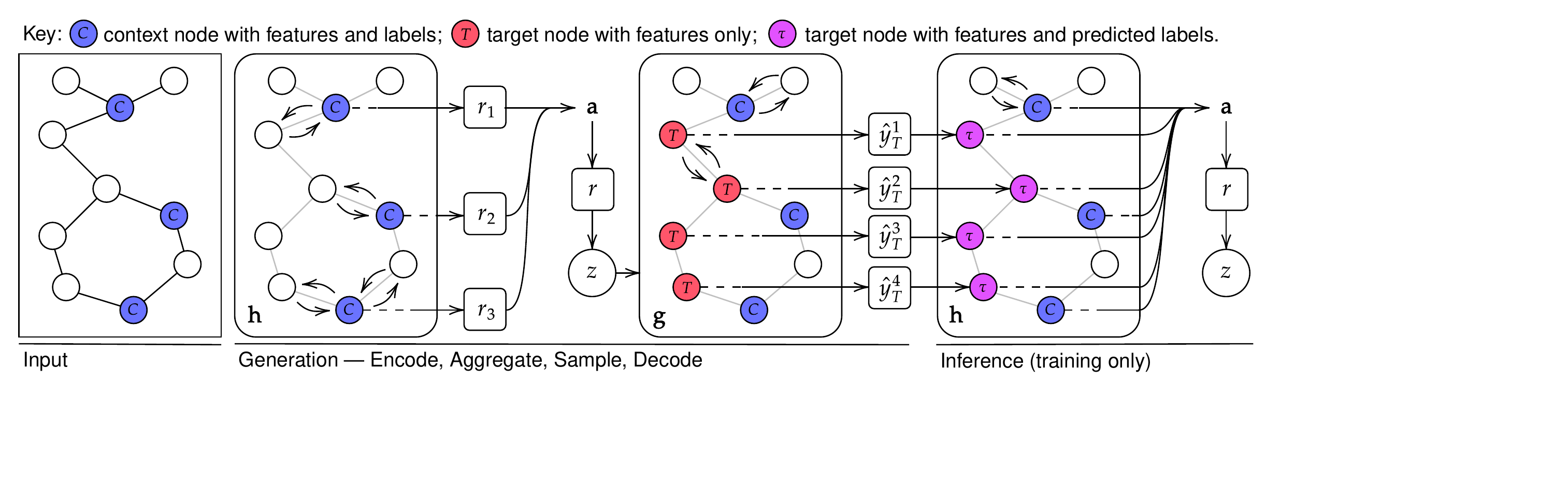}
    \caption{Computational graph of the Message Passing Neural Process. \textbf{Input:} the dataset consists of examples (nodes) and a relational structure (edges). Features, $x$, are observed for every node, but labels are only observed for the context set, the {\color{figblue}blue nodes} labelled $C$. \textbf{Generation:} the encoder, $\mathbf{h}$, uses message-passing operations over the dataset to produce neighbourhood-aware representations of the context set, $r_i$. The aggregator, $\mathbf{a}$, combines these into a single representation, $r$, which parameterises the global latent variable, $z$. The decoder, $\mathbf{g}$, which also uses message-passing operations, is conditioned on a sample from the global latent variable and makes label predictions over the {\color{figred}target set}, $\hat{y}_T$. \textbf{Inference:} the predicted labels are added to the target examples, differentiated from the unlabelled targets by the label $\uptau$ and {\color{figpurple}purple nodes}. The  dataset is again passed through the encoder, $\mathbf{h}$, and aggregator, $\mathbf{a}$, to produce the global latent variable as conditioned on the joint target and context set, as required in the ELBO objective (Equation \ref{eq:mpnpelbo}) for training. Note: most message-passing arrows have been omitted for clarity.}
    \label{fig:mpnp}
\end{figure*}

We present Message Passing Neural Processes (MPNPs) as the synthesis of the MP and NP models. Figure \ref{fig:mpnp} illustrates the operation of an MPNP. We describe each step below (the Appendix contains pseudocode for the entire computation).

\subsubsection{Problem Statement} Given a partially-labelled set of \emph{nodes} with features $\mathbf{X}$ \emph{and neighbours given by $\mathbf{A}$}, sampled from $f : \mathbf{X, A} \rightarrow \mathbf{Y}$, with $f \sim \mathcal{D}$, the goal is to predict labels for a subset of the unlabelled \emph{nodes}.

\subsubsection{Dataset Sampling} In the classification setting, the context set for a dataset (here, a \emph{graph}) is defined as a set $\mathcal{C} = \big\{(\mathbf{x}_i, \mathbf{y}_i)\big\}$ of nodes and their one-hot labels. The information available to the encoder $h$ is given by the set $\mathcal{C} \cup \big\{\mathbf{x}_j~|~j \in \bigcup_{i \in \text{context set}} \EuScript{N}(i)\big\}$, with $|\mathcal{C}| = m$, which contains the context set and the $k$-hop neighbourhoods of all context nodes. In this way, the MPNP uses the relational structure between the context set and other nodes to produce richer representations of the context nodes. In turn, the global latent variable $\mathbf{z}$ is able to encode relational structure present in the underlying stochastic process. The target set, $\mathcal{T} = \{\mathbf{x}_i~|~i \in \text{context set}\} \cup \{\mathbf{x}_i~|~i \notin \text{context set}\}$, with $|\mathcal{T}| = n$, is a superset of the context set (though not necessarily containing the entire graph), \emph{without labels}. The decoder $g$ also uses information from the $k$-hop neighbourhood when predicting target labels.

\subsubsection{Encoder} The encoder $h$ takes as input elements from the context set, encoded as $\mathbf{h}_i = \mathbf{x}_i \parallel \mathbf{y}_i$, along with node features from their $k$-hop neighbourhoods, where $\parallel$ denotes concatenation. Zero-vectors are used in place of labels for nodes outside the context set. A representation is produced for every element of the context set, $\mathbf{r}_i$, using $T$ message-passing operations, as defined in Equation \ref{eq:mp}.

\subsubsection{Aggregation} The representations $\mathbf{r}_i$ for all context nodes are aggregated into a single vector $\mathbf{r} = \mathbf{a}(\{\mathbf{r}_i\})$ via a permutation-invariant function $\mathbf{a}$, as shown in Figure \ref{fig:mpnp}.
A normal distribution $\mathbf{z}~\sim~\mathcal{N}(\mu_z(\mathbf{r}), \mathrm{diag}[\sigma_z(\mathbf{r})])$ is assumed over the global latent variable $\mathbf{z}$, where $\mu_z$ and $\sigma_z$ are linear transformations of $\mathbf{r}$.

\subsubsection{Decoder} The input to the decoder $g$ is the concatenation of a sample $\mathbf{z}'$ from this distribution with each of the target feature vectors, i.e. $\mathbf{h}'_i = \mathbf{x}_i \parallel \mathbf{z}'$. This step calculates label predictions $\hat{\mathbf{y}}_i$ for the nodes in $\mathcal{T}$ in a similar manner to the encoding step, producing the output $\mathbf{r}'_i$.
Following evaluation convention established in \citet{le2018empirical}, the target label predictions are $\hat{\mathbf{y}}_i \sim \mathcal{N}\big(\text{softmax}(\mu_y(\mathbf{r}'_i)), \mathrm{diag}[0.1 + 0.9 \times \text{softplus}(\sigma_y(\mathbf{r}'_i))]\big)$, with linear transformations $\mu_y,\sigma_y$.

\subsubsection{Generation and Inference} Starting from Equation~\ref{eq:npgen}, with the function $\gamma$ corresponding to the neural network $\mathbf{g}$ in Figure~\ref{fig:mpnp}, and letting $\mathbf{x}_{\EuScript{N}(i)}$ denote features corresponding to an entire neighbourhood, we state the generative model for the MPNP (the Appendix contains a complete derivation):
\begin{multline}\label{eq:mpnpgen}
    p(\mathbf{z}, \mathbf{y}_{1:n}~|~\mathbf{x}_{1:n}, \bigcup_{i=1}^n \mathbf{x}_{\EuScript{N}(i)}) = \\ p(\mathbf{z}) \prod_{i=1}^n \EuScript{N}\big(\mathbf{y}_i~|~F(\mathbf{x}_i \| \mathbf{z}, \square {}_{j \in \EuScript{N}(i)}, G(\mathbf{x}_i \| \mathbf{z}, \mathbf{x}_j \| \mathbf{z})), \sigma^2\big).
\end{multline}
The decoder function $g$ is a composition of learnable functions (linear projections, MP steps) and non-linearities, so it is trainable with amortised variational inference. The variational posterior $q(\mathbf{z} | \mathbf{x}_{1:n}, \mathbf{y}_{1:n})$ is also parameterised by a neural network ($\mathbf{h}$ in Figure~\ref{fig:mpnp}) that is permutation-invariant, as each of the functions in $\mathbf{h}$ satisfies this property (full proof in the Appendix). Optimisation can be achieved using standard methods with the ELBO objective (fully derived in the Appendix), where $D = \mathbf{x}_{1:n} \cup \bigcup_{i=1}^n \mathbf{x}_{\EuScript{N}(i)} \cup \mathbf{y}_{1:n}$:
\begin{multline} \label{eq:mpnpelbo}
    \log p\big(\mathbf{y}_{m+1:n}~|~\mathbf{x}_{1:n}, \bigcup_{i=1}^n \mathbf{x}_{\EuScript{N}(i)}, \mathbf{y}_{1:m}\big) \geq 
    \\ \sum_{i=m+1}^n \mathbb{E}_{q(\mathbf{z}|D)} \big[\log p(\mathbf{y}_i~|~\mathbf{x}_i, \mathbf{x}_{\EuScript{N}(i)}, \mathbf{z})\big] \\
    - ~\mathbb{K}\mathbb{L}\Big(q(\mathbf{z}~|~\mathbf{x}_{1:n}, \bigcup_{j=1}^n \mathbf{x}_{\EuScript{N}(j)}, \mathbf{y}_{1:n}) \\
    \Big\Vert q(\mathbf{z}~|~\mathbf{x}_{1:m}, \bigcup_{j=1}^m \mathbf{x}_{\EuScript{N}(j)}, \mathbf{y}_{1:m})\Big).
\end{multline}

\subsubsection{Aggregation in Challenging Settings} The manner in which information is stored in the global latent variable $\mathbf{z}$ is crucial---at test time, the (context-conditioned) sample is processed together with the new target points, so it must reflect the behaviour of the new stochastic process in a way that is relevant to the task. Despite a simple mean over $r_i$ being sufficient for many tasks, it is often necessary to produce a class-aware representation. Therefore, we adopt the alternative aggregation function used by~\citet{garnelo2018conditional} for few-shot learning tasks,
\begin{equation}
    a'(\{\mathbf{r}_i\}) \triangleq \Big\Vert_{\mathbf{c} \in C} a(\mathbb{I}_{\{\mathbf{c}\}}(\mathbf{y}_i) * \mathbf{r}_i),
\end{equation}
where $C$ is the set of classes \emph{in the current context}, with $|C|$ fixed, as required. This performs concatenation ($\parallel$) of per-class summaries aggregated with $a$. Intuitively, different classes in the context set are clearly delimited in this scheme, which is especially helpful in few-shot learning settings, where novel classes are seen during testing. Models using this scheme have the `-c' suffix.

\section{Experiments} \label{s:tasks}

\subsection{Baselines and model details}
We evaluate against a variety of baselines that collectively leverage all sources of information present in the tasks (featural, relational \& contextual). This helps highlight where the advantages of the MPNP lie in a given setting. The \textbf{label propagation algorithm} (LP) \citep{ZhuLearningPropagation} makes direct use of the context points provided at test time (nodes are labelled by their neighbours, who label their neighbours, and so on) and is best suited to segmentation-like tasks. Where relevant, we include \textbf{guessing the most common context-label} (Mode), as this may significantly outperform the uniform-prior ($\sfrac{1}{N}$) for some tasks. \textbf{Graph neural networks} (GNNs) use training data in the inductive setup, but not the additional context points provided at test time. They are expected to perform well on tasks with fixed classes and little variation in the generative process across the set of datasets being modelled. We note that these models are not designed to handle arbitrary labelling tasks and their expected performance is bound by chance, i.e. $\mathrm{E}[\mathrm{acc.}]=\sfrac{1}{N}$: as predictions do not depend on class labellings, for any given task example we can construct a set of equivalent tasks by permuting the labels, and over the set of permutations the average performance will be chance (a formal derivation is provided in the Appendix). As such, we do not include this baseline on such tasks. In our setup, the GNN consists of GCN layers with skip-connections (the Appendix contains a detailed description).

Non-message-passing \textbf{Neural Processes} (NPs) are limited only by their inability to leverage relational information between points, though this is, of course, a serious limitation in the settings we consider. We use the same \textbf{Message Passing Neural Process} and NP architectures for most Cora, ShapeNet and biochemical tasks, with the addition of Maxout layers \cite{Goodfellow2013MaxoutNetworks} for CA tasks. Other modifications are described with the experiment in which they are used, and full model details for each scenario are provided in the Appendix.

\subsection{Fixed labelling tasks}
We first consider tasks where the same set of classes appear in every example and the class labelling is `fixed'. Inductive GNNs are designed for this setting and provide a useful baseline performance.

Two tasks are adapted from the \textbf{TUD collection}~\cite{KKMMN2016}: Enzymes and DHFR.
The Enzymes dataset consists of proteins represented as networks of secondary-structural elements ($\alpha$-helices, $\beta$-sheets, $\beta$-turns; SSEs) with biochemical features describing these units and edges between connected elements.
DHFR is a library of small molecules that inhibit a particular protein, represented as graphs of atoms connected by bonds with spatial positions as features.
Table \ref{t:node-tu} shows the MPNP narrowly outperforms the NP at the Enzymes task and by a much greater margin for DHFR, though in each case an inductive GNN is more successful.
This suggests that the relational information present in the Enzymes dataset is of secondary importance to the featural information of the SSEs, and that there is limited variation over both datasets, given that an inductive model can perform well without any context points. Nevertheless, it is promising that the MPNP is able to use the relational information in DHFR to improve greatly on the NP.

The \textbf{Protein-Protein Interaction (PPI) Site Prediction} task involves predicting which nodes (amino acids) in an amino acid residue graph are involved in an experimentally-determined PPI \citep{Zeng2019}.
Solving this task is thought to depend strongly on being able to use relational information, and there is great variation between examples.
As expected following the TUD results, the MPNP excels in this setting, with SOTA-competitive results at plausible context rates presented in Table~\ref{t:ppi}.
The prefix `R' indicates that the message-passing scheme has an edge-type dependency, as in the R-GCN \cite{schlichtkrull2018modeling}. Full details for this model are provided in the Appendix. 

\begin{table}
  \scriptsize
  \caption{Node classification on biochemical datasets. Accuracy reported at $\{5, 10, 30\}$\% context points. {\ubold first} / {\usecond second}.}
  \label{t:node-tu}
  \centering
  \begin{tabular}{lccc|ccc}
    \toprule
    & \multicolumn{3}{c}{Enzymes} & \multicolumn{3}{c}{DHFR}\\
    Model & 5 & 10 & 30 & 5 & 10 & 30\\
        \midrule
    NP       & \usecond79.23 &93.43&\usecond95.75 & 54.66&55.71&57.38 \\
    MPNP     & 79.09&\usecond94.10&\ubold95.78 & \usecond 88.65 & \usecond 89.62 & \usecond90.53 \\
    \midrule
    GNN      & \ubold94.23 &\ubold 94.23 & 94.23&\ubold 93.35 &\ubold 93.35 &\ubold 93.35 \\
    LP & 58.93&63.91&76.42 & 38.48& 41.51&53.63\\
    \bottomrule
  \end{tabular}
\end{table}

\begin{table}
    \scriptsize
    \caption{Node classification on Protein-Protein Interaction Site Prediction. R-MPNP scores for $\{5,30\}$\% sampling rates. Results for ISIS, DeepPPISP and R-GCN are taken from \citet{Ofran2007}, \citet{Zeng2019}, and \citet{schlichtkrull2018modeling}, respectively.}
    \label{t:ppi}
    \centering
    \begin{tabular}{lcc|cc|cc}
    \toprule
    Method & \multicolumn{2}{c}{Accuracy \%} & \multicolumn{2}{c}{F-measure} & \multicolumn{2}{c}{MCC} \\
    \midrule
    ISIS & \multicolumn{2}{c|}{69.4} & \multicolumn{2}{c|}{0.267} & \multicolumn{2}{c}{0.097} \\
    DeepPPISP  & \multicolumn{2}{c|}{65.5} & \multicolumn{2}{c|}{\ubold 0.397} & \multicolumn{2}{c}{0.206} \\
    R-GCN & \multicolumn{2}{c}{76.7} & \multicolumn{2}{c}{0.165} & \multicolumn{2}{c}{0.169} \\
     \cmidrule(lr){2-7}
     & 5 & 30 & 5 & 30 & 5 & 30 \\
     \cmidrule(lr){2-7}
     NP & 77.5 & 79.3 & 0.212 & 0.180 & 0.145 & 0.150 \\
     R-MPNP & \ubold 79.1 & \ubold 80.7 & 0.292& 0.348  & \ubold 0.236 & \ubold     0.284 \\
     \bottomrule
    \end{tabular}
\end{table}

\begin{figure*}[t]
    \begin{center}
    \includegraphics[trim=5 5 5 5, clip, width=\textwidth]{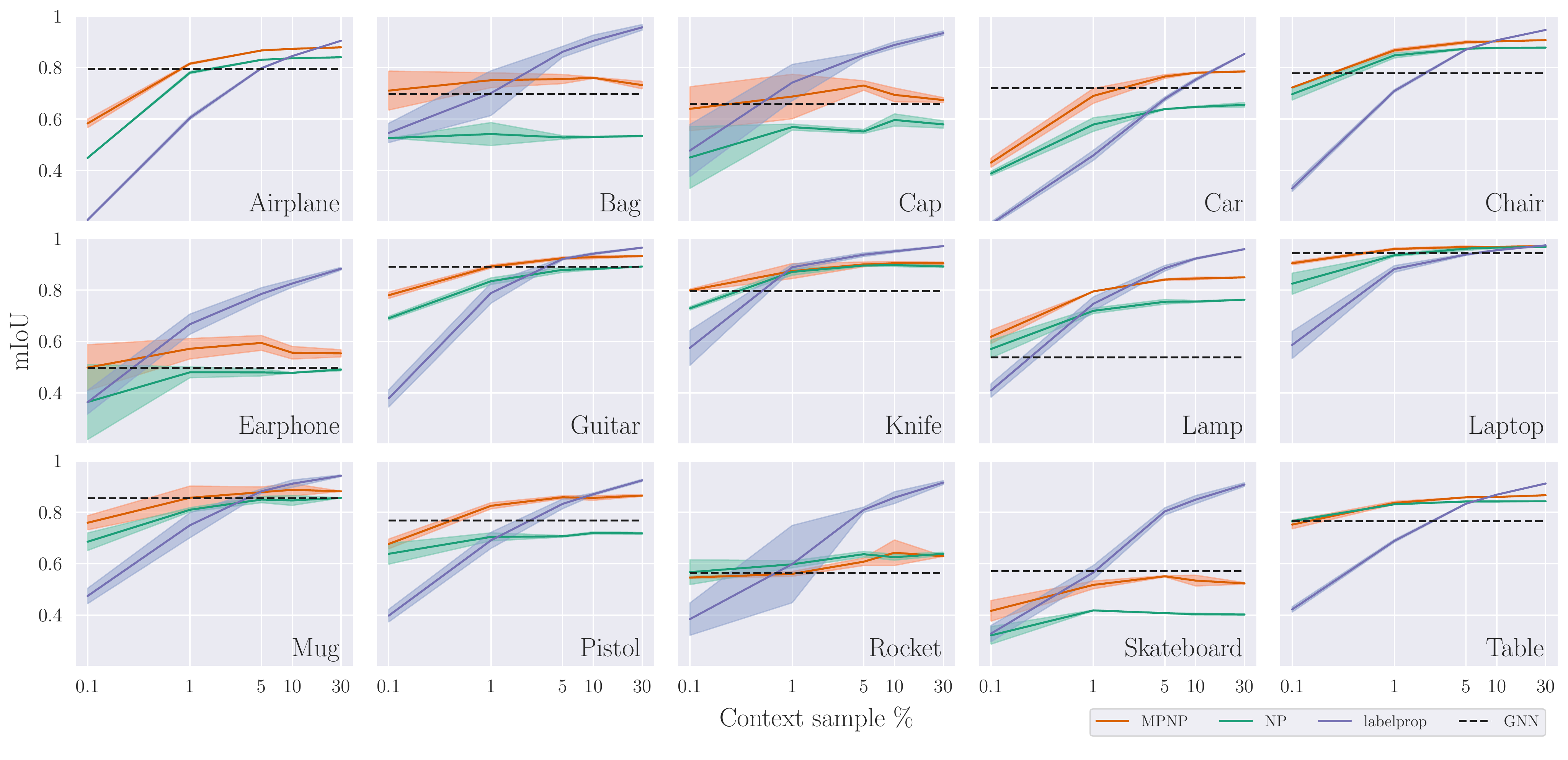}
    \end{center}
    \caption{Linear-log plots of mIoU over context sample rates with 95\% confidence interval shading for the fixed-class ShapeNet task, by category. The GNN is inductive and does not depend on context sampling. Numerical results are given in the Appendix.}
    \label{fig:shapenet_grid_plots}
    \centering
    \includegraphics[trim=15 5 5 25, clip, width=\textwidth]{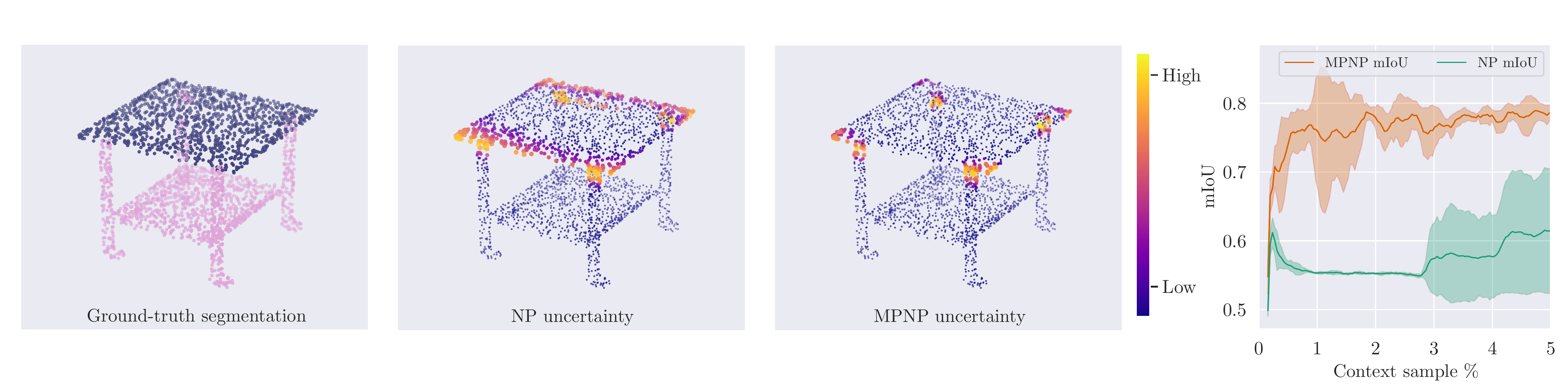}
    \caption{Segmentation uncertainty over an example from the ShapeNet fixed-class \emph{table} category test set and active sampling. \emph{(Left:)} Ground truth labels are shown for the table-top (purple) and table-leg (pink) parts. \emph{(Centre:)} Uncertainty is depicted by the size and colour of the points: higher at larger, yellower points and lower at smaller, bluer points. \emph{(Right:)} Active sampling.}
    \label{fig:shapenet_uncertainty}
\end{figure*}

The \textbf{ShapeNet} repository \cite{Chang2015ShapeNet:Repository,YiACollections} is a collection of large-scale 3D shapes, represented as point clouds for our applications.\footnote{There exist many techniques that make fuller use of the geometric information available, but for this proof-of-concept we consider only the simplest method.} We embed the points as a nearest-neighbours graph ($\mathbf{A}$) and use the ($x,y,z$) position as node features ($\mathbf{X}$). There are 16 object categories, each one having a fixed number of parts, ranging from two to six. The labels have consistent meaning across datasets within a category. For example, we model the process that produces \textit{chairs} with \textit{arms, legs, seats} and \textit{backs}, which we can consistently label $\{1,2,3,4\}$.

\textbf{Part labelling} results are presented in Figure~\ref{fig:shapenet_grid_plots}. We use the mean-Intersection-over-Union (mIoU) metric, which is standard for segmentation tasks: the ratio of overlap (TP) to the union (TP+FP+FN) is found for each part, and averaged (higher is better, T/F P/N = true/false positives/negatives). In 11 object categories, the MPNP outperforms the NP at more than 95\% confidence across the entire context sampling range, and is the top-performing model over some of the sampling range in 13 out of 15 categories. At 30\% sampling, label propagation dominates as expected.

Figure~\ref{fig:shapenet_uncertainty} shows the superior uncertainty-modelling capabilities of the MPNP. In the first 3 plots, we \textbf{visualise the uncertainty predictions} for a \emph{table} sample. Though the models achieve similar mIoU, the MPNP is only significantly uncertain at the borders between parts (a physically relevant uncertainty), whereas the NP is uncertain along the table-top edges, which are distant from any table-leg points in the internal geometry of the table. On the right, we present the results of an \textbf{active learning experiment} similar to that described by~\citet{garnelo2018conditional}. At each step, the target with the greatest uncertainty is added to the context set (i.e. labelled) and predictions are repeated. This shows the power of useful uncertainty estimates in the MPNP.

\subsubsection{Cellular Automata} Irregular graph-CAs have been used to study traffic networks~\cite{Malecki2017GraphSimulation}, social networks~\cite{Hunt2011UsingBehavior}, urban and regional development~\cite{White1998CitiesAutomata,OSullivan2001Graph-CellularModel} and logistics~\cite{Lopez2019MicroscopicZone}, and cell dynamics~\cite{Bock2010GeneralizedDynamics}. Our aim is to show how MPNPs extend the available model capabilities, as existing baselines are likely to struggle. The model is provided with the states of some cells over a generation and tasked with evolving others. To evaluate generalisation, we prevent rule-set overlap in the train, validation and test sets. This contrasts with the existing work of~\citet{Gilpin2018CellularNetworks}, where the model learns a single rule-set, and that of \citet{mordvintsev2020growing}, who train a CA to produce a desired pattern. We provide an overview of these tasks, with full details given in the Appendix.
\begin{figure}[h]
    \begin{center}
    \includegraphics[trim = 980 210 500 190, clip, width=0.9\columnwidth]{{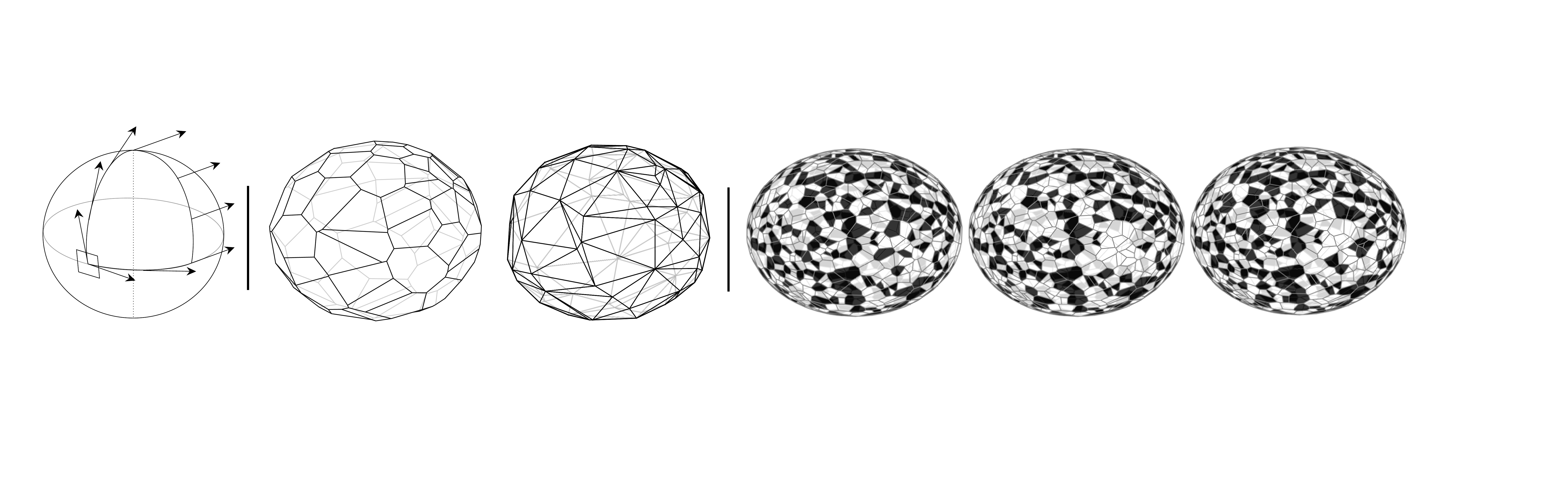}}
    \end{center}
    \caption{Two generations of a population-density CA on a spherical Voronoi network producing complex patterns in the cells. The MPNP receives the first state (nodes are cells, edges link bordering cells, features are $0/1$ according to cell state) and predicts cell states ($0/1$) after one transition.}
    \label{fig:CA}
    \vspace{-10pt}
\end{figure}

Conway's Game of Life~\cite{games1970fantastic} consists of cells in a 2D lattice governed by simple rules: cells become alive/are born (B) or stay alive/survive (S) depending on the number of living neighbours. The \textbf{Life-like family} of CA are the generalisations of these rules over any number of neighbours 0--8, defining $2^{18}$ variants. Neighbour counts can also be generalised to neighbourhood population-densities, and \textbf{density-based rules} can be adapted to irregular graphs and non-planar topologies. We consider single-interval rules, such that cells live or die based on being inside or outside a continuous range of population-densities, on small-world, scale-free, and spherical Voronoi networks (an example of the latter is shown in Figure \ref{fig:CA}).
\begin{figure}[h]
    \centering
    \includegraphics[trim=25 20 25 10, clip, width=0.95\columnwidth]{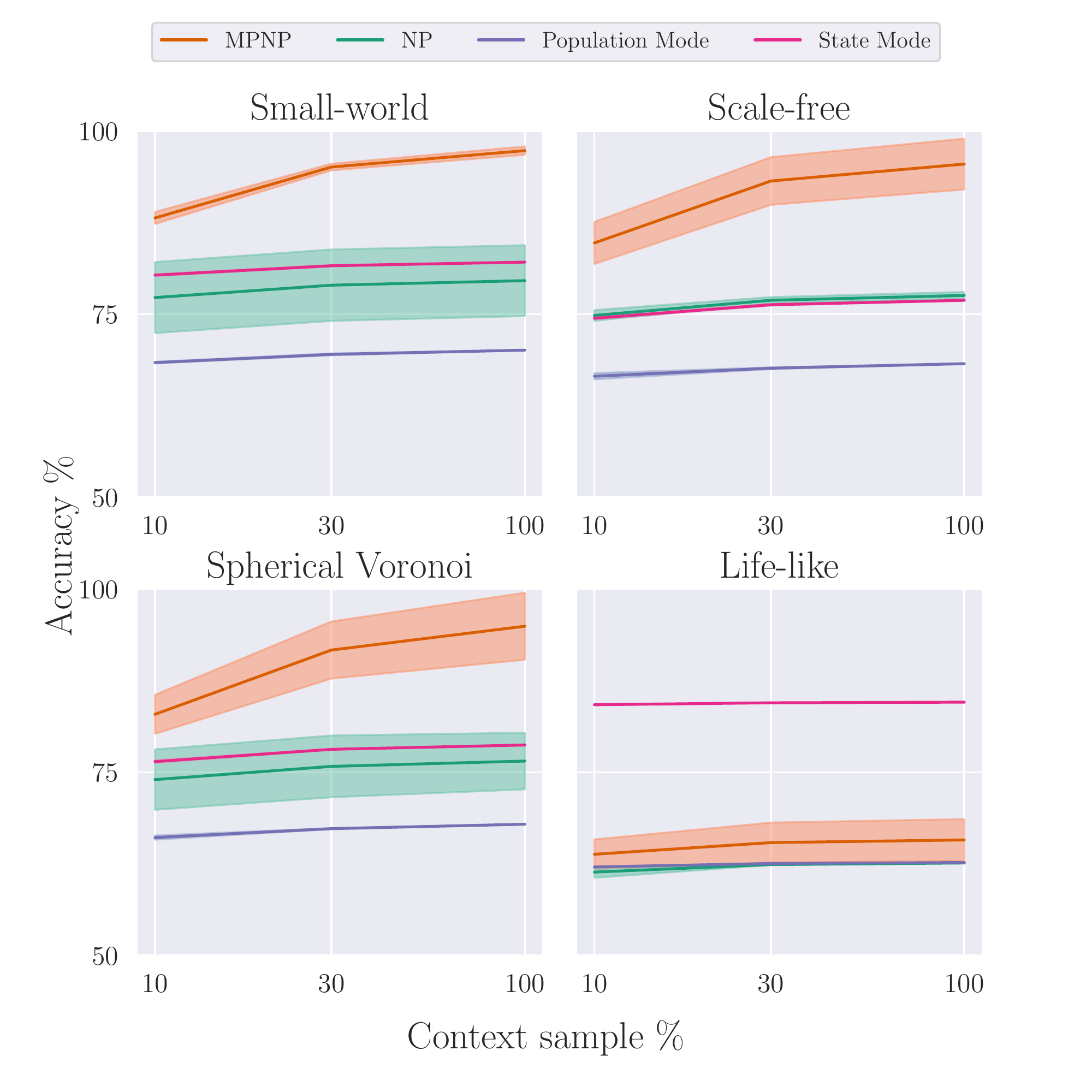}
    \caption{State evolution accuracy $\pm \sigma$ for density- and count-based cellular automata. Models are trained at 30-50\% context sampling. Testing at 100\% effectively judges the quality of the rule embedding under perfect information.}
    \label{fig:CA_grid}
\end{figure}

State evolution results are presented in Figure~\ref{fig:CA_grid}. Here, the ‘Population/State Mode’ baselines are versions of ‘guessing-the-mode‘ that output the most common label over the whole context set or by initial state, respectively. The NP is often able to match the state-mode strategy, but this is the ceiling to methods that do not take relational information into account. The MPNP is able to learn effective representations that generalise well to the disjoint test set for density-based rules. For density-based rules, MPNPs perform strongly across a variety of graph structures, while NPs are bound by simple strategies that guess the most common state change. Neither model is able to perform well for the Life-like family, despite the existence of a solution to this problem for MPNPs, outlined in the Appendix.

\subsection{Arbitrary labelling tasks}
\citet{garnelo2018conditional} applied the CNP model in the arbitrary labelling setting, where each dataset includes samples drawn from a fixed number $k$ of class types, where the total number of types $K \gg k$. As the total number of classes could be very large and test examples may include unseen classes, using fixed-classes is infeasible. Instead, arbitrary labellings $(1,...,k)$ are assigned on a per-dataset basis, and models are required to adapt accordingly.

The Cora-ML task is a widely used community detection benchmark. Papers are represented by bag-of-words vectors with edges indicating that one of the papers cited the other. Our task, \textbf{Cora-Branched}, is derived from the less popular but more complete dataset, with 70 classes over 11 computer science disciplines \cite{McCallum2000AutomatingLearning,BojchevskiDEEPRANKING}. There are ten times as many classes and the bag-of-words feature vectors are tripled in length. Given a partially labelled subgraph of the network, the task is to label the rest. We consider the \textbf{transductive setup} \cite{Yang2016RevisitingEmbeddings,Velickovic2017GraphNetworks} where every class is observed during training and as a \textbf{few-shot learning task} where, at test time, the models are presented with classes entirely unobserved during training. Results for the transductive setting are presented in Table~\ref{tab:cora_branched}. Both models perform well in the low-sampling rate regime, indicating a strong feature signal, though the MPNP-c significantly outperforms the NP-c in every test, by up to 10\% for 7-class. LP performs best at higher sampling rates and for more classes, as expected. Table~\ref{tab:fsl_cora} compares the quality of NP and MPNP representations in the few-shot learning context---the MPNP is better able to generalise to unseen categories.

\begin{table}[h]
    \centering
        \scriptsize
        \caption{Results on the Cora-Branched transductive learning tasks for 3, 7 and 11 classes (\#). Mean accuracy and standard deviations are reported at $\{1, 5, 10, 30\}$\% context points.}
        \label{tab:cora_branched}
        \sisetup{%
            table-align-uncertainty=true,
            separate-uncertainty=true,
            detect-weight=true,
            detect-inline-weight=math,
            mode=text
        }
        \begin{tabular}{p{1ex}l|*{4}{S[table-format=2.2]@{\,\(\tinypm \)\,}S[table-format=1.2]}}
            \toprule
            \# & Model & \multicolumn{2}{c}{1\%} & \multicolumn{2}{c}{5\%} & \multicolumn{2}{c}{10\%} & \multicolumn{2}{c}{30\%} \\
            \midrule \multirow{4}{*}{3}
            & NP-c      & 67.00 & 1.83 & 76.99 & 1.50 & 78.56 & 1.19 & 79.61 & 1.20 \\
            & MPNP-c    & \ubold 79.71 & \ubold 1.04 & \ubold 88.28 & \ubold 0.59  & \ubold 89.41 & \ubold 0.58 & \ubold 90.02 & \ubold 0.60 \\
            & LP & 65.31 & 0.73 & 75.57 & 0.31 & 77.90 & 0.16 & 82.04& 0.18 \\
            & Mode      & 54.35 & 0.10 & 54.28 & 0.07 & 54.40 & 0.27 & 54.41 & 0.18 \\
            \midrule \multirow{4}{*}{7}
            & NP-c    & 52.83 & 0.49 & 63.02 & 0.50  & 64.29 & 0.43 & 65.23 & 0.51   \\
            & MPNP-c   & \ubold 58.40 & \ubold 0.77 & \ubold 68.96& \ubold1.08&\ubold 70.53& \ubold 0.88 & 71.54& 0.91\\
            & LP & 52.62 & 0.31 & 64.85& 0.22& 68.55& 0.14& \ubold74.96 & \ubold0.20 \\
            & Mode & 30.48 & 0.16 & 30.57 & 0.07 & 30.50 & 0.10 & 30.50 & 0.10 
            \\
            \midrule \multirow{4}{*}{11}
            & NP-c     & 34.57 & 2.18 & 37.94 & 0.84 & 38.88 & 0.80 & 39.42 & 0.78 \\
            & MPNP-c  & 43.62& 1.01 & 50.64& 1.14& 51.87& 1.23& 52.67& 1.24\\
            & LP & \ubold 46.84 & \ubold 0.55 & \ubold 60.11& \ubold 0.12 & \ubold 64.22 & \ubold 0.08& \ubold71.73 & \ubold0.05 \\
            & Mode & 21.60 & 0.08 & 21.60 & 0.11 & 21.63 & 0.09& 21.66 & 0.10
            \\
            \bottomrule 
        \end{tabular}
\end{table}

\begin{table}[h]
  \scriptsize
  \caption{Performance on the Cora-Branched few-shot learning tasks for 2, 3, 5 and 11 class ($\#$) tasks. Accuracy at $\{1,5,10\}$\% context points.}
  \label{tab:fsl_cora}
  \centering
  \begin{tabular}{clccc}
    \toprule
    \# & Model & 1\% & 5\% & 10\% \\
    \midrule
    \multirow{2}{*}{2}
    & NP-c   & 59.25 & 63.53 & 64.29 \\
    & MPNP-c & \ubold 62.91 & \ubold 67.53 & \ubold 68.57 \\
    \midrule
    \multirow{2}{*}{3}
    & NP-c   & 49.82 & 56.93 & 59.03 \\
    & MPNP-c & \ubold 53.83 & \ubold 63.75 & \ubold 64.52 \\
    \midrule
    \multirow{2}{*}{5}
    & NP-c   & 36.84 & 42.68 & 44.10 \\
    & MPNP-c & \ubold 41.67 & \ubold 49.99 & \ubold 51.15 \\
    \midrule
    \multirow{2}{*}{11}
    & NP-c   & 19.71 & 21.13 & 21.82 \\
    & MPNP-c & \ubold 23.56 & \ubold 26.00 & \ubold 27.44 \\
    \bottomrule
  \end{tabular}
\end{table}

In the \textbf{ShapeNet mixed-category} setup, we model the process that produces $n$-part objects (say, $n=4$ for \textit{chairs} with \textit{arms, legs, seats} and \textit{backs}, as well as \textit{airplanes} with \textit{engines, bodies, tails} and \textit{wings}.) Here, labels have consistent meaning only within a given realisation, so using a fixed ordering of labels implies a meaningless relationship between, say, \textit{chair-backs} and \textit{airplane-wings}. We thus provide an arbitrary permutation of class labels for each example.

\begin{table}
    \scriptsize
    \centering
    \begin{tabular}{clcccc}
        \toprule
        \# & Model & 0.1\% & 1\% & 5\% & 10\% \\
        \midrule \multirow{4}{*}{2}
         &	 NP-c	& 48.06 & 83.60 & 88.62 & 89.17 \\
         & MPNP-c   & \ubold 57.18 & \ubold 86.08 & 90.81 & 91.37 \\
         & LP       & 55.55 & 84.37 & \ubold 91.90 & \ubold 93.93 \\
         & GNN      & 36.14 & 36.14 & 36.14 & 36.14 \\
        \midrule \multirow{4}{*}{3}
         &	 NP-c	& \ubold 46.87 & 76.66 & 81.12 & 81.47 \\
         & MPNP-c   & 45.52 & \ubold 78.95 & 83.80 & 84.31 \\
         & LP       & 41.12 & 69.84 & \ubold 84.40 & \ubold 87.76 \\
         & GNN      & 21.68 & 21.68 & 21.68 & 21.68 \\
        \midrule \multirow{4}{*}{4}
         &	 NP-c	& 28.48 & 67.19 & 72.30 & 72.88 \\
         & MPNP-c   & \ubold 31.52 & \ubold 74.30 & 81.38 & 82.20 \\
         & LP       & 30.29 & 66.61 & \ubold 83.61 & \ubold 87.91 \\
         & GNN      & 15.82 & 15.82 & 15.82 & 15.82 \\
        \bottomrule
    \end{tabular}
    \caption{ShapeNet mixed-category, arbitrary-labelling results for 2, 3, and 4-part shapes ($\#$). We report the mIoU for $\{0.1, 1, 5, 10\}$\% context points.}
    \label{t:node-shapenetmixed}
\end{table}

Table~\ref{t:node-shapenetmixed} shows results for the mixed-class part-grouped ShapeNet task. The GNN struggles as expected, with performance below chance. Label propagation is the strongest performer at high sampling rates, with the MPNP-c and NP-c at a relative advantage with fewer context points. The NP-c performs best at 0.1\% on 3-class, which may be due to category imbalances (80\% of 3-part objects are tables) disrupting the MPNP-c. MPNP-c and label propagation otherwise divide the sampling range as top performers.

\subsection{Discussion} The results presented show that the richer context representations and structural bias of the MPNP are generally beneficial, outperforming the NP on Cora-Branched, PPISP, 4/5 TUD tasks, ShapeNet mixed (excluding 3-class@0.1\%), while producing semantically-realistic uncertainties, as shown in Figure~\ref{fig:shapenet_uncertainty} and Figure~\ref{fig:more-shapenet-viz} in the Appendix. Label propagation is more successful when more labels are available, but MPNP vastly improves on it at low sampling rates, showing powerful capabilities in scarce data settings. GNNs learn better when the generative process has little functional variation, but perform poorly in the opposite case (mixed-class and few-shot), and are entirely unsuitable in the arbitrary labelling setting. The TUD biochemical datasets are the only fixed-class setting where GNNs do consistently better than MPNPs, though we can attribute this to the lack of functional variation of the generative process in these narrow tasks. On ShapeNet and PPISP fixed-class tasks, MPNP surpasses the GNN in most cases.

\section{Conclusion}

We have introduced the Message Passing Neural Process, an NP model that leverages the explicit structure between samples from a stochastic process for classification. Our work supplies NPs with the inductive bias necessary to model the relational structure in each dataset, similarly to the ConvCNP model that adds the translation equivariance inductive bias. Therefore, the data points are represented in a context-aware manner, rather than an isolated one. The stronger representations obtained achieve notable performance improvements in few-shot learning and rule-based settings, while uncertainty estimates become more meaningful with respect to the dataset structure. In future work, we will incorporate attention in the MPNP (similarly to Attentive NPs) and aim to model the structural generative process, to allow sampling entire graphs from the latent variable.

\section{Ethics Statement}

The primary group to benefit from our work would be machine learning researchers developing graph representation learning and uncertainty-oriented methods. Since the results we have presented in Table~\ref{t:ppi} are competitive with state-of-the-art methods, biochemistry researchers and practitioners may wish to investigate further uses of the MPNP. The cellular automata (CA) community might also adopt MPNPs in their research. The wide applicability of CAs might thus be inherited by our model.

Whilst there are no groups that are obviously immediately disadvantaged by this work, we can imagine a scenario in which deploying a model like our own on social networks could technologically enable or enhance certain repressive policies or undermine democratic institutions. For example, the use of ads in social media to target individuals with (mis)information is well established as having played a critical role in the 2016 US election and 2015 UK EU referendum. These methods may be made cheaper as a result of better uncertainty modelling over networks. Though it is unknown whether greater access to these technologies (from the reduction in cost) will restore balance in democracies, in a repressive regime it is more than likely the case that this application would only serve to strengthen the state. None of our applications explore this use case (predicting behaviour or beliefs of individuals in a social network) and we would caution against such research.

While we believe that our contribution is more foundational in nature, it may nevertheless suffer from general limitations of machine learning algorithms. Similarly to most ML setups, we have trained MPNPs on data from certain distributions. Due to the explicit uncertainty modelling present in the MPNP, out-of-distribution samples might produce (potentially wrong) predictions with higher uncertainty estimates---in some cases, these greater uncertainty values could indicate that the model output might be biased by the data it was trained on. In these kinds of situations, we believe that incorporating uncertainty is a strength of our model, providing additional signal to the user when compared to a standard deep learning method that only outputs the predicted class.

Should the work be taken up by biochemistry researchers, the application is likely to be computational exploration of molecules for desirable properties (e.g. druggability). As our approach is essentially model-free (besides the inductive biases associated with using a graph representation) there may be less reason to expect strong generalisation and so, despite the improvements seen for the test set, using the model for out-of-distribution exploration may be more prone to failure than simpler, mechanistic models. Here the consequence of system failure or bias is likely limited to wasted resources in the wet-lab as there is no suggestion that the results of these models should be used to directly produce and apply drugs without existing safety protocols (nor is it common practice to do so.)

\section{Acknowledgements}
The authors would like to thank Ramon Vi{\~n}as Torn{\'e}, Nikola Simidjievski and Cristian Bodnar for their helpful comments, and Duo Wang, Felix Opolka, Jacob Deasy, Emma Rocheteau, Conor Sheehan, Penelope Jones, Petar Veli{\v{c}}kovi{\'c}\ and Toby Shevlane for their comments on an earlier version of this work.
We also acknowledge the following tools used for experiments and presenting our work: Figma, Mathcha, Desmos, Overleaf, Weights and Biases and the Color Brewer project.

\bibliographystyle{aaai}
\bibliography{extrabib}

\clearpage
\begin{appendices}

\section{MPNP Details}

\subsection{Generative Model} \label{a:mpnp_gen}

Equation~\ref{eq:npgen} lets us derive the MPNP generative model, where the function $\gamma$ corresponds to the neural network $g$ in Figure~\ref{fig:mpnp} and $\mathbf{x}_{\EuScript{N}(i)}$ denotes the features corresponding to the neighbourhood of node $i$:
\begin{equation}
\begin{aligned}
    &p(\mathbf{z}, \mathbf{y}_{1:n}~|~\mathbf{x}_{1:n}, \bigcup_{i=1}^n \mathbf{x}_{\EuScript{N}(i)}) =\\
    &p(\mathbf{z}) \prod_{i=1}^n p(\mathbf{y}_i~|~\mathbf{x}_i, \mathbf{x}_{\EuScript{N}(i)}, \mathbf{z}) =\\
    &p(\mathbf{z}) \prod_{i=1}^n \EuScript{N}\big(\mathbf{y}_i~|~\gamma(\mathbf{x}_i, \mathbf{x}_{\EuScript{N}(i)}, \mathbf{z}), \sigma^2\big) =\\
    &p(\mathbf{z}) \prod_{i=1}^n \EuScript{N}\big(\mathbf{y}_i~|~F(\mathbf{x}_i \| \mathbf{z}, \bigodot_{j \in \EuScript{N}(i)}, G(\mathbf{x}_i \| \mathbf{z}, \mathbf{x}_j \| \mathbf{z})), \sigma^2\big).
\end{aligned}
\end{equation}
In this derivation, line 2 assumes that $p(\mathbf{y}_i~|~\mathbf{x}_i, \mathbf{x}_{\EuScript{N}(i)}, \mathbf{z})$ takes the form of a normal distribution, with mean and variance being functions of $\mathbf{x}_i, \mathbf{x}_{\EuScript{N}(i)}, \mathbf{z}$. Line 3 uses the fact that, in our model, $\gamma = \text{ReLU} \circ L_2 \circ \textit{MP}^T \circ \text{ReLU} \circ L_1$. Let us first consider the case for $T = 1$. The function $G$ corresponds to a linear transformation $L_1 = \mathbf{W}_{\text{MP}}$ applied to each of the (target node) neighbours' feature vectors (here, we refer to the concatenated representations $\mathbf{x}_j \| \mathbf{z}$). This is followed by leveraging the aggregation operator $\bigodot_{j \in \EuScript{N}(i)}$ within the neighbourhood of each target node. Finally, $F$ consists of applying the skip-connection (linear transformation) $\mathbf{W}_{\text{skip}}$ to each of the target node feature vectors, followed by the ReLU activation of the MP step and $\text{ReLU} \circ L_2$. The only difference for $T = 2$ lies in the aggregator and linear transformations within the MP step being performed twice. It is important to note that the variance $\sigma^2$ is output by the same network $\gamma$, as each prediction has its own associated uncertainty.

\subsection{Model Pseudocode} \label{a:mpnp_pseudo}

Algorithm 1 summarises the MPNP label generation process described in the \textbf{Message Passing Neural Processes} section.

\begin{algorithm}
\SetKwInOut{Input}{Input}
\SetKwInOut{Output}{Output}
\Input{Context set $\mathcal{C} = \{\mathbf{x}_i, \mathbf{y}_i\}$, with $|\mathcal{C}| = m$, features of context set node neighbours $\{\mathbf{x}_i~\Vert~j \in \bigcup_{i \in \text{context set}} \EuScript{N}(i)\}$, target set $\mathcal{T} = \{\mathbf{x}_i~|~i \in \text{context set}\} \cup \{\mathbf{x}_i~|~i \notin \text{context set}\}$, with $|\mathcal{T}| = n > m$, features of target set node neighbours $\{\mathbf{x}_i~\Vert~j \in \bigcup_{i \in \text{target set}} \EuScript{N}(i)\}$.}

\Output{Target label predictions $\{\hat{\mathbf{y}}_i~\Vert~i \in \text{target set}\}$.}

\tcp{Initialise node features}
\ForEach{$i \in$ context set}{
    $\mathbf{h}^0_i \leftarrow \mathbf{x}_i \parallel \mathbf{y}_i$\\
    \ForEach{$j \in \bigcup_{i \in \text{context set}} \EuScript{N}(i)$}{
        $\mathbf{h}^0_j \leftarrow \mathbf{x}_j \parallel \mathbf{0}$\\
    }
}
\tcp{Encoding}
\ForEach{$i \in$ context set}{
    $\mathbf{h}^0_i \leftarrow \text{ReLU}(L_1(\mathbf{h}^0_i))$\\
    \ForEach{$j \in \bigcup_{i \in \text{context set}} \EuScript{N}(i)$}{
        $\mathbf{h}^0_j \leftarrow \text{ReLU}(L_1(\mathbf{h}^0_j))$\\
    }
}
\ForEach{$t \in {1, ..., T}$}{
    \ForEach{$i \in$ context set}{
        $\mathbf{h}^{t}_i \leftarrow \textit{MP}(\mathbf{h}^{t-1})$
    }
}
\ForEach{$i \in$ context set}{
    $\mathbf{r}_i \leftarrow L_2(\mathbf{h}^T_i)$
}
\tcp{Aggregation}
$\mathbf{r} \leftarrow a(\{\mathbf{r}_i~\Vert~i \in \text{context set}\})$\\
\tcp{Decoding}
Sample $\mathbf{z}'~\sim~\mathcal{N}(\mu(\mathbf{r}), \mathrm{diag}[\sigma(\mathbf{r})])$\\
\ForEach{$i \in$ target set}{
    $\mathbf{h}'^0_i = \mathbf{x}_i \parallel \mathbf{z}'$\\
    \ForEach{$j \in \bigcup_{i \in \text{target set}} \EuScript{N}(i)$}{
        $\mathbf{h}'^0_j \leftarrow \mathbf{x}_j \parallel \mathbf{z}'$\\
    }
}
\ForEach{$i \in$ target set}{
    $\mathbf{h}'^0_i \leftarrow \text{ReLU}(L_1(\mathbf{h}'^0_i))$\\
    \ForEach{$j \in \bigcup_{i \in \text{target set}} \EuScript{N}(i)$}{
        $\mathbf{h}'^0_j \leftarrow \text{ReLU}(L_1(\mathbf{h}'^0_j))$\\
    }
}
\ForEach{$t \in {1, ..., T}$}{
    \ForEach{$i \in$ target set}{
        $\mathbf{h}'^{t}_i \leftarrow \textit{MP}(\mathbf{h}'^{t-1})$
    }
}
\ForEach{$i \in$ target set}{
    $\mathbf{r}'_i \leftarrow \text{ReLU}(L_2(\mathbf{h}'^T_i))$\\
    $\hat{\mathbf{y}}'_i \sim \mathcal{N}\Big(\text{softmax}(\mu(\mathbf{r}'_i))$,\\ ~~~~~~~~~~~~~~~~$\mathrm{diag}[\big(0.1 + 0.9 \times \text{softplus}(\sigma(\mathbf{r}'_i))]\big)\Big)$
}
\caption{MPNP computation.}
\label{algo:mpnp}
\end{algorithm}

\subsection{Encoder Permutation Invariance} \label{a:mpnp_perminv}
We show that, for initial node representations $\mathbf{h}_i$, the transformation $\mathbf{r}_i = (L_2 \circ \textit{MP}^T \circ \text{ReLU} \circ L_1) (\mathbf{h}_i)$ produced by the encoder is permutation-invariant:
\begin{equation}
\begin{aligned}
    &\forall~\text{permutation}~\mathbf{\Pi}.\\
    &(L_2 \circ \textit{MP}^T \circ \text{ReLU} \circ L_1) (\mathbf{X\Pi}, \mathbf{\Pi^TA\Pi}) =\\
    &\big((L_2 \circ \textit{MP} \circ \text{ReLU} \circ L_1) (\mathbf{X}, \mathbf{A})\big)\mathbf{\Pi}.
\end{aligned}
\end{equation}
\emph{Proof:} Assume an arbitrary set of features $\mathbf{X} \in \mathbb{R}^{n \times d}$ and an adjacency matrix $\mathbf{A} \in \{0,1\}^{n \times n}$, where $n$ is the number of nodes in the context set and $d$ is the feature dimensionality. We first show that each of the operations within the encoder is permutation-invariant:
\begin{enumerate}
    \item The linear projections $L_1, L_2$ are applied to each of the node vectors $\mathbf{X}_i$ separately, so changing the order of input nodes will result in the same order in the output:
    \begin{equation}
    \begin{aligned}
        L_i(\mathbf{X\Pi}, \mathbf{\Pi^TA\Pi})
        &= L_i(\mathbf{X\Pi}), \forall i \in \{1,2\} \\
        &= (L_i(\mathbf{X}_{\Pi_1})~L_i(\mathbf{X}_{\Pi_2})~\dots~L_i(\mathbf{X}_{\Pi_n}))^\text{T} \\
        &= (L_i(\mathbf{X}_1)~L_i(\mathbf{X}_2)~\dots~L_i(\mathbf{X}_n))^\text{T}\mathbf{\Pi} \\
        &= L_i(\mathbf{X})\mathbf{\Pi} \\
        &= L_i(\mathbf{X}, \mathbf{A})\mathbf{\Pi}.\qed
    \end{aligned}
    \end{equation}
    \item The same holds for the activation functions, which are applied element-wise:
    \begin{equation}
    \begin{aligned}
        \text{ReLU}(\mathbf{X\Pi}, \mathbf{\Pi^TA\Pi})
        &= \text{ReLU}(\mathbf{X\Pi}) \\
        &= \text{ReLU}(X_{\Pi_ij}), \forall i, j \\
        &= \text{ReLU}(X_{ij})\mathbf{\Pi} \\
        &= \text{ReLU}(\mathbf{X})\mathbf{\Pi} \\
        &= \text{ReLU}(\mathbf{X}, \mathbf{A})\mathbf{\Pi}.\qed
    \end{aligned}
    \end{equation}
    \item The message passing operation is also permutation-invariant, since the transformation $\mathbf{A} \rightarrow \mathbf{P^TAP}$ preserves the structure of the graph, with node neighbourhoods undergoing the transformation $\EuScript{N}(i) \triangleq \{j~|~A_{ij} = 1\} \rightarrow \EuScript{N}(i)\mathbf\Pi \triangleq \{\Pi_j~|~A_{\Pi_i\Pi_j} = 1\}$:
    \begin{equation}
    \begin{aligned}
        &\textit{MP}(\mathbf{X\Pi}, \mathbf{\Pi^TA\Pi})\\
        &=\text{ReLU}\big(\mathbf{W}_{\text{skip}}(\mathbf{X\Pi})_i+\sum_{j' \in \EuScript{N}(i)\mathbf\Pi} \mathbf{W}_{\text{MP}}(\mathbf{X\Pi})_j\big),\\
        &\text{where}~j' = \Pi_{j},\\
        &=\text{ReLU}\big(\mathbf{W}_{\text{skip}}(\mathbf{X}_{\Pi_i})~+\sum_{\Pi_j \in \EuScript{N}(\Pi_i)} \mathbf{W}_{\text{MP}}(\mathbf{X}_{\Pi_j})\big) \\
        &=\text{ReLU}\big((\mathbf{W}_{\text{skip}}\mathbf{X}_i)\mathbf\Pi~+\sum_{j \in \EuScript{N}(i)} (\mathbf{W}_{\text{MP}}\mathbf{X}_j)\mathbf\Pi\big) \\
        &=\text{ReLU}\big(\mathbf{W}_{\text{skip}}\mathbf{X}_i~+\sum_{j \in \EuScript{N}(i)} \mathbf{W}_{\text{MP}}\mathbf{X}_j\big)\mathbf\Pi \\
        &=\textit{MP}(\mathbf{X}, \mathbf{A})\mathbf\Pi.\qed
    \end{aligned}
    \end{equation}
\end{enumerate}
Each type of operation performed within the encoder is thus permutation-invariant. Composing permutation-invariant functions yields a function which has this property itself, so it follows that the overall transformation is permutation-invariant.$\qed$

\subsection{ELBO} \label{a:mpnp_elbo}
We derive the ELBO objective stated under \textbf{Generation and Inference}. In the derivation, we assume $m$ context nodes and $n$ target nodes (that is, $n - m$ additional targets). The aim is to maximise the log-likelihood of target labels $\mathbf{y}_{m+1:n}$, given the target node features $\mathbf{x}_{1:n}$, context node features $\mathbf{x}_{1:m}$, context labels $\mathbf{y}_{1:m}$ and neighbourhoods of context nodes. We denote by $\mathbf{x}_{\EuScript{N}(i)}$ the features corresponding to an entire neighbourhood and let $D = \mathbf{x}_{1:n} \cup \bigcup_{i=1}^n \mathbf{x}_{\EuScript{N}(i)} \cup \mathbf{y}_{1:n}$.
\begingroup
\allowdisplaybreaks
\begin{align*}
    &\log p\big(\mathbf{y}_{m+1:n}~|~\mathbf{x}_{1:n}, \bigcup_{i=1}^n \mathbf{x}_{\EuScript{N}(i)}, \mathbf{y}_{1:m}\big) =\\
    &\log p\big(\mathbf{y}_{m+1:n}, \mathbf{z}~|~\mathbf{x}_{1:n}, \bigcup_{i=1}^n \mathbf{x}_{\EuScript{N}(i)}, \mathbf{y}_{1:m}\big) -\\
    &\log p\big(\mathbf{z}~|~\mathbf{x}_{1:n}, \bigcup_{i=1}^n \mathbf{x}_{\EuScript{N}(i)}, \mathbf{y}_{1:n}\big) =\\
    &\Big[\log p\big(\mathbf{z}~|~\mathbf{x}_{1:m}, \bigcup_{i=1}^m \mathbf{x}_{\EuScript{N}(i)}, \mathbf{y}_{1:m}\big) + \sum_{i=m+1}^n \log p(\mathbf{y}_i~|~\mathbf{x}_i, \mathbf{x}_{\EuScript{N}(i)}, \mathbf{z})\Big]~-\\
    & ~~~\log p\big(\mathbf{z}~|~\mathbf{x}_{1:n}, \bigcup_{i=1}^n \mathbf{x}_{\EuScript{N}(i)}, \mathbf{y}_{1:n}\big) =\\
    &\log \frac{p\big(\mathbf{z}~|~\mathbf{x}_{1:m}, \bigcup_{i=1}^m \mathbf{x}_{\EuScript{N}(i)}, \mathbf{y}_{1:m}\big)}{q\big(\mathbf{z}~|~\mathbf{x}_{1:n}, \bigcup_{i=1}^n \mathbf{x}_{\EuScript{N}(i)}, \mathbf{y}_{1:n}\big)} + \sum_{i=m+1}^n \log p(\mathbf{y}_i~|~\mathbf{x}_i, \mathbf{x}_{\EuScript{N}(i)}, \mathbf{z})~-\\
    &\log \frac{p\big(\mathbf{z}~|~\mathbf{x}_{1:n}, \bigcup_{i=1}^n \mathbf{x}_{\EuScript{N}(i)}, \mathbf{y}_{1:n}\big)}{q\big(\mathbf{z}~|~\mathbf{x}_{1:n}, \bigcup_{i=1}^n \mathbf{x}_{\EuScript{N}(i)}, \mathbf{y}_{1:n}\big)} =\\
    &\mathbb{E}_{q(\mathbf{z}|D)} \Bigg[\sum_{i=1}^n \log p(\mathbf{y}_i~|~\mathbf{x}_i, \mathbf{x}_{\EuScript{N}(i)}, \mathbf{z}) +\\
    & ~~~~~~~~~~~~~~~\log \frac{p(\mathbf{z}~|~\mathbf{x}_{1:m}, \bigcup_{j=1}^m \mathbf{x}_{\EuScript{N}(j)}, \mathbf{y}_{1:m})}{q(\mathbf{z}~|~\mathbf{x}_{1:n}, \bigcup_{j=1}^n \mathbf{x}_{\EuScript{N}(j)}, \mathbf{y}_{1:n})}\Bigg] +\\
    &\mathbb{K}\mathbb{L}\Big(q(\mathbf{z}~|~\mathbf{x}_{1:n}, \bigcup_{j=1}^n \mathbf{x}_{\EuScript{N}(j)}, \mathbf{y}_{1:n})\Big\Vert p(\mathbf{z}~|~\mathbf{x}_{1:n}, \bigcup_{j=1}^n \mathbf{x}_{\EuScript{N}(j)}, \mathbf{y}_{1:n})\Big) \geq\\
    &\mathbb{E}_{q(\mathbf{z}|D)} \Bigg[\sum_{i=1}^n \log p(\mathbf{y}_i~|~\mathbf{x}_i, \mathbf{x}_{\EuScript{N}(i)}, \mathbf{z}) +\\
    & ~~~~~~~~~~~~~~~\log \frac{p(\mathbf{z}~|~\mathbf{x}_{1:m}, \bigcup_{j=1}^m \mathbf{x}_{\EuScript{N}(j)}, \mathbf{y}_{1:m})}{q(\mathbf{z}~|~\mathbf{x}_{1:n}, \bigcup_{j=1}^n \mathbf{x}_{\EuScript{N}(j)}, \mathbf{y}_{1:n})}\Bigg] =\\
    &\sum_{i=m+1}^n \mathbb{E}_{q(\mathbf{z}|D)} \big[\log p(\mathbf{y}_i~|~\mathbf{x}_i, \mathbf{x}_{\EuScript{N}(i)}, \mathbf{z})\big] -\\
    &\mathbb{E}_{q(\mathbf{z}|D)} \log \frac{q(\mathbf{z}~|~\mathbf{x}_{1:n}, \bigcup_{j=1}^n \mathbf{x}_{\EuScript{N}(j)}, \mathbf{y}_{1:n})}{p(\mathbf{z}~|~\mathbf{x}_{1:m}, \bigcup_{j=1}^m \mathbf{x}_{\EuScript{N}(j)}, \mathbf{y}_{1:m})} =\\
    &\sum_{i=m+1}^n \mathbb{E}_{q(\mathbf{z}|D)} \big[\log p(\mathbf{y}_i~|~\mathbf{x}_i, \mathbf{x}_{\EuScript{N}(i)}, \mathbf{z})\big] -\\
    & ~~~\mathbb{K}\mathbb{L}\Big(q(\mathbf{z}~|~\mathbf{x}_{1:n}, \bigcup_{j=1}^n \mathbf{x}_{\EuScript{N}(j)}, \mathbf{y}_{1:n})\Big\Vert q(\mathbf{z}~|~\mathbf{x}_{1:m}, \bigcup_{j=1}^m \mathbf{x}_{\EuScript{N}(j)}, \mathbf{y}_{1:m})\Big).
\end{align*}
\endgroup
In the order given above, the (in)equalities use the following: rewriting the log-likelihood via the posterior distribution, substituting the first term via the generative model, introducing a variational distribution $q(\mathbf{z}~|~\mathbf{x}_{1:n}, \bigcup_{i=1}^n \mathbf{x}_{\EuScript{N}(i)}, \mathbf{y}_{1:n})$ (in our case, the encoder $h$ and the aggregation $a$) to approximate the posterior $p(\mathbf{z}~|~\mathbf{x}_{1:n}, \mathbf{y}_{1:n})$, multiplying by $q\big(\mathbf{z}~|~\mathbf{x}_{1:n}, \bigcup_{i=1}^n \mathbf{x}_{\EuScript{N}(i)}, \mathbf{y}_{1:n}\big)$ and integrating over $\mathbf{z}$, the result that $\forall p, q.~\mathbb{K}\mathbb{L}(p\|q) \geq 0$, separating terms, approximating $p(\mathbf{z}~|~\mathbf{x}_{1:m}, \bigcup_{j=1}^m \mathbf{x}_{\EuScript{N}(j)}, \mathbf{y}_{1:m})$ with $q(\mathbf{z}~|~\mathbf{x}_{1:m}, \bigcup_{j=1}^m \mathbf{x}_{\EuScript{N}(j)}, \mathbf{y}_{1:m})$ and applying the $\mathbb{K}\mathbb{L}$ definition.

\section{Task Descriptions}
\label{apdx:task_descriptions}

\begin{table*}[h]
    \centering
    \caption{Dataset statistics by tasks. For transductive Cora there is a single citation network (i.e. 1 graph) from which subgraphs are sampled to produce training examples (of which the total possible number depends on the number of classes being used in the split e.g. for the 2-class task there are ${11 \choose 2} = 55$.) In the few-shot case, the train and test subgraphs are disjoint and neither features, labels, nor edges are observed from the test set during training. In all Cora tasks we use PCA to reduce the number of input features from 8710 to 100. For Proteins, we remove 6 graphs with more than one component or non-physical features (negative length). The density CA tasks (Voronoi, spherical Voronoi, small-world, scale-free) use generated graphs with the number of nodes being drawn from $[100, 200]$, we report the observed mean as generated by our seed.}
    \begin{tabular}{lccccr}
    \toprule
         Dataset & Task & Graphs & Mean-Nodes & Features & Classes \\
         \midrule
         \multirow{19}{*}{ShapeNet} & Bag & 76 & 2749.46 & 3 & 2 \\
         & Cap & 55 & 2631.53 & 3 & 2 \\
         & Knife & 392 & 2156.57 & 3 & 2 \\
         & Laptop & 451 &2758.13 & 3 & 2 \\
         & Mug & 184 & 2816.97 & 3 & 2 \\
         \cmidrule{2-6}
         & \emph{2-parts} & 1158 & 2,557.26 & 3 & 2 \\
         \cmidrule{2-6}
         & Earphone & 69 & 2496.70 & 3 & 3 \\
         & Guitar & 787 & 2353.91 & 3 & 3 \\
         & Pistol & 283 & 2654.22 & 3 & 3 \\
         & Rocket & 66 & 2358.59 & 3 & 3 \\
         & Skateboard & 152 &2529.55 & 3 & 3 \\
         & Table & 5271 & 2722.40 & 3 & 3 \\
         \cmidrule{2-6}
         & \emph{3-parts} & 6628 & 2,665.34 & 3 & 3 \\
         \cmidrule{2-6}
         & Airplane & 2690 & 2577.92 & 3 & 4 \\
         & Car & 898 & 2763.81 & 3 & 4 \\
         & Chair & 3758 & 2705.34 & 3 & 4 \\
         & Lamp & 1547 & 2198.46 & 3 & 4 \\
         \cmidrule{2-6}
         & \emph{4-parts} & 8893 & 2,584.53 & 3 & 4 \\
         \cmidrule{2-6}
         & Motorbike & 202 & 2735.65& 3 & 6 \\
         \midrule
         \multirow{5}{*}{TUD} & Proteins & 1113 & 39.06 & 29 & 3 \\
         & Enzymes & 600 & 32.63 & 18 & 3 \\
         & DHFR & 467 & 42.23 & 3 & 9 \\
         & COX2 & 467 & 41.22 & 3 & 8 \\
         & BZR & 405 & 35.75 & 3 & 10 \\
         \midrule
         PPISP & & 408 & 207.64 & 38 & 2\\
         \midrule
         \multirow{2}{*}{Cora} & Transductive & 1 & 19,793 & 100* & 70 \\
         & Few-shot train & 1 & 17,657 & 100* & 11 \\
         & Few-shot test & 1 & 2136 & 100* & 11 \\
         \midrule
         \multirow{5}{*}{Cellular Automata} & Life-like & 2659 & 900 & 2 & 2 \\
         & Voronoi & 2700 & 149.81 & 2 & 2\\
         & Spherical-Voronoi & 2700 & 150 & 2 & 2 \\
         & Small-world (WS) & 2700 & 149.54 & 2 & 2 \\
         & Scale-free (BA) & 2700 & 149.34 & 2 & 2\\
         \bottomrule
    \end{tabular}
    \label{tab:my_label}
\end{table*}

\begin{table*}[t]
    \centering
    \caption{Class-ID information for the Cora class taxonomy. There are 11 disciplines collectively containing 70 classes. These IDs can be used to select classes as loaded by \texttt{CitationFull} from PyTorch Geometric.}
    \begin{tabular}{lcr}
    \toprule
         Discipline & \texttt{IDs}  \\
        \midrule
         Information Retrieval & \{0,1,4,12\} \\
         Databases & \{2,10,28,42,44,46,60\} \\
         Artificial Intelligence & \{5,8,9,11,14,22,33,34,48,53,54\} \\
         Machine Learning & \{3,20,29,55,57,58,59\} \\
         Encryption and Compression & \{6,15,26\} \\
         Operating Systems & \{7,27,45,62\} \\
         Networking & \{13,16,24,30\} \\
         Hardware and Architecture & \{17,40,41,50,67,68,69\} \\
         Data-Structures Algorithms and Theory & \{18,19,21,31,32,35,61,64,66\} \\
         Programming & \{23,36,37,49,51,52,56,63,65\} \\
         Human Computer Interaction & \{25,38,39,43,47\}\\
         \bottomrule
         \end{tabular}
    \label{tab:cora_class_ids}
\end{table*}

\paragraph{Cellular Automata} For the Life-like family of cellular automata we sample $\sim\%$ of the possible $2^{18}$ rule sets at random (Bernoulli $p=0.01$). For each selected rule set, we generate a random state on a $30\times30$ toroidal lattice (top connects to bottom, left connects to right) and check that every possible state is present (i.e. there are live cells with each of 0, 1, 2, ..., 8 neighbours and similarly a dead cell), then step forward one generation by applying the rule set to form the input-label pair. For density-based rules we use birth/survival functions with either the form of the top-hat function:
\begin{align}
    R_0(d, k_1, k_2) =
    \begin{cases}
        0 & \textup{for } d < k_1, \\
        1 & \textup{for } k_1 \leq d \leq k_2, \\
        0 & \textup{for } d > k_2.
    \end{cases}
\end{align}
or $1-R_0$, i.e.:
\begin{align}
    R_1(d, k_1, k_2) =
    \begin{cases}
        1 & \textup{for } d < k_1, \\
        0 & \textup{for } k_1 \leq d \leq k_2, \\
        1 & \textup{for } d > k_2.
    \end{cases}
\end{align}
The irregular graphs that the density-based rules operate on are generated using Scipy and NetworkX. In each case we sample the number of nodes uniformly from the interval $[100, 200]$. For the planar Voronoi the nodes are positioned at uniformly at random in the unit square and the tessellation is generated using SciPy.\footnote{\url{https://docs.scipy.org/doc/scipy/reference/generated/scipy.spatial.Voronoi.html}} For spherical-Voronoi the nodes are positioned uniformly at random over the surface of the sphere and the tessellation is generated using SciPy.\footnote{\url{https://docs.scipy.org/doc/scipy/reference/generated/scipy.spatial.SphericalVoronoi.html}} For small-world the graphs are generated using the Watts-Strogatz model with $p=0.1$ and $k=10$ i.e. the network is initialised in a ring-lattice connected to its 10 nearest-neighbours on the ring and then edges are rerouted with probability $0.1$, using the NetworkX implementation.\footnote{\url{https://networkx.github.io/documentation/networkx-1.9/reference/generated/networkx.generators.random_graphs.watts_strogatz_graph.html}} For the scale-free case we use the Barabasi-Albert model with $m=3$, using the NetworkX implementation.\footnote{\url{https://networkx.github.io/documentation/networkx-1.9/reference/generated/networkx.generators.random_graphs.barabasi_albert_graph.html}}

\paragraph{Cora} We base our Cora tasks on the \texttt{CitationFull} dataset provided in PyTorch Geometric\footnote{\url{https://pytorch-geometric.readthedocs.io/en/latest/_modules/torch_geometric/datasets/citation_full.html}} which is loading the data used by Bojchevski and G{\"u}nnemann\footnote{\url{https://github.com/abojchevski/graph2gauss}}, who in turn base their set on that originally gathered by Andrew McCallum of University of Massachussets Amherst.\footnote{\url{https://people.cs.umass.edu/~mccallum/data.html}} Nodes are research papers with bag-of-word features (8710 words meet the threshold for inclusion by Bojchevski and G{\"u}nnemann, which was given through correspondence with the authors as a minimum of appearing in 10 documents in the set) that use presence/absence rather than counts (multi-hot). Edges indicate that one of the papers cited the other, though we do not distinguish between citing/cited and the graph is undirected. The papers belong to one of 70 topics within 11 disciplines of Computer Science, and we present the relevant class-ID information for splitting by discipline in Table \ref{tab:cora_class_ids}. In the few-shot learning setup we separate out classes $\{29,4,10,53,26,45,30,17,21,56,47\}$ for validation and $\{59,1,42,48,15,62,16,67,61,49,38\}$ for testing, representing 14.94\% and 10.79\% of the total nodes, respectively. Each of these splits contains a class from every branch (hence 11 classes) with an effort made to ensure the class-to-branch ratios were also approximately 15\% and 10\%, with preferential selection for the test set and an allowance for producing largely connected subgraphs. For example, on the Encryption-branch there are three classes 15, 26, and 6, containing approximately one sixth, one third and one half of the nodes, respectively, with class 15 being selected for the test set and 26 for the validation set. Practically the connectivity allowance means selecting class 48 (12.1\% of AI) rather than 22 (9.0\% of AI) for the test set and class 29 (14.7\% of ML) rather than 55 (14.9\% of ML) for the validation set.

\paragraph{TUD Datasets} Proteins and Enzymes are more commonly treated as graph-classification tasks, but there is an intermediate labelling of secondary structural elements ($\alpha$-helices, $\beta$-sheets and $\beta$-turns) that can be used in the node classification setup. DFHR, COX2 and BZR consist of small libraries of small molecule inhibitors against each respective protein target (Dihydrofolate Reductase, Cycloxygenase-2 and the Benzodiazapene Receptor). In the typical graph-classification task, molecules are deemed active or inactive on the basis of a thresholded half-maximal inhibitory concentration measure determined through \textit{in vitro} biochemical assays. The node-classification task considered here requires the model to predict node labels representing encodings of atom-type. Node features are \textit{xyz} coordinates of the conformation provided in the datasets. 

\paragraph{Protein-Protein Interaction Site Prediction} This node-classification task utilises protein structural data collated in \citep{Zeng2019}, representing protein structures as graphs of interacting residues. Nodes are featurised with low dimensional embeddings of physicochemical properties \citep{Meiler2001}, encodings of secondary structure, solvent accessibility metrics, and position-specific scoring matrices which capture evolutionary information as protein-protein interaction residues have been shown to be evolutionarily conserved. Edge features represent one-hot encodings of intramolecular interaction types. Node labels indicate whether or not that amino acid takes part in an experimentally determined protein-protein interaction. Graphs are constructed using graphein.\footnote{\url{https://github.com/a-r-j/graphein}}

\section{Experimental and Model Details}
\label{apdx:exp_model_details}

All models were trained on a Titan Xp GPU or an RTX 2080 GPU, with \texttt{torch.manual\_seed(0)} across all experiments. An 80/20 train/test split was used for TUD datasets\footnote{\url{https://pytorch-geometric.readthedocs.io/en/latest/modules/datasets.html#torch_geometric.datasets.TUDataset}} and the ones provided by PyTorch Geometric\footnote{\url{https://pytorch-geometric.readthedocs.io/en/latest/modules/datasets.html#torch_geometric.datasets.ShapeNet}} for all ShapeNet tasks. \textbf{The supplementary material includes code for all models and experiments described in this paper.}

\subsection{MPNP}

The architecture of the MPNP can be summarised as follows:
\begin{enumerate}
    \item encoder: Linear($h$), ReLU, \{MP($h$), ReLU\}$ \times T$, Linear($r$);
    \item global latent variable encoder: Linear($r$), [Linear($z$), Linear($z$)] (mean \& variance of $\mathbf{z}$);
    \item decoder: Linear($h$), ReLU, \{MP($h$), ReLU\}$ \times T$, Linear($h$), ReLU, [Linear($C$), Linear($C$)] (mean \& variance of $\mathbf{\hat{y}}$).
\end{enumerate}

Across all experiments, the Adam optimiser is used to maximise the ELBO (i.e. minimise the sum of the negative log-likelihood and KL-divergence in equation~\ref{eq:mpnpelbo}).

\subsubsection{TUD} On Proteins and Enzymes, the MPNP hyperparameters are $h = 64$, $r = 128$, $z = 256$; for the MPNP-c, $h = 64$, $r = 96$, $z = 288$; both have $T = 2$. On DHFR, COX2 and BZR, both MPNP and MPNP-c have $h = 64$, $r = 128$, $z = 256$, $T = 1$. We trained both models for 400 epochs with learning rate $7e\times10^{-5}$ on all datasets except for Enzymes, where we used 700 epochs and learning rate $1\times10^{-4}$. For all datasets, we sample context and (additional) target points in the $10\%$--$25\%$ range.

\subsubsection{ShapeNet} Across all experiments, $h = 64$, $r = 128$, $z = 256$, $T = 2$. The MPNP was trained for 400 epochs on fixed-class and 500 epochs on mixed-class tasks, with $5\%$--$25\%$ context and (additional) target points and a learning rate of $7\times10^{-5}$.

\subsubsection{Cora} In both the transductive and few-shot settings, $h = 64$, $r = 64$, $T = 2$ and $z = N\times64$ for $N$-classes. In the transductive setting the model is trained for 500 epochs where little if any overfitting is observed. In the few-shot setting the model is trained for 400 epochs. The model performs significantly better on the training classes in the few-shot case, though this is expected. A learning rate of $7\times10^{-5}$ is used in both cases and we sample context and target points in the $10\%$--$50\%$ range.

\subsubsection{CA} The CA models use a modified architecture that includes Maxout layers \cite{Goodfellow2013MaxoutNetworks} that can be summarised as follows:
\begin{enumerate}
    \item encoder: MP($h$), ReLU, \{Linear($h$), ReLU\}$\times 3$, Maxout($h,2$), Linear($r$), ReLU;
    \item global latent variable encoder: Linear($r$), [Linear($z$), Linear($z$)] (mean \& variance of $\mathbf{z}$);
    \item decoder: MP($h$), ReLU, \{Linear($h$), ReLU\}$\times 3$, Maxout($h,2$), (concatenation with $\mathbf{z}$), \{Linear($h$), ReLU\}$\times 3$, [Linear($C$), Linear($C$)] (mean \& variance of $\mathbf{\hat{y}}$).
\end{enumerate}
Maxout layers use a pool-size of 2 and the decoder delays concatenation with $\mathbf{z}$ until after the Maxout layer (and the part before concatenation matches the encoder). For both the life-like and density-based settings, $h=64$, $r=64$, $z=128$. The models are trained for 200 epochs with a learning rate of $1\times10^{-4}$ and we sample context and target points in the $30\%$--$50\%$ range.

\subsubsection{R-MPNP}
The architecture of the R-MPNP can be summarised as follows:
\begin{enumerate}
    \item encoder: Linear($h$), ReLU, \{R-MP($h$), ReLU\}$ \times T$, Linear($r$);
    \item global latent variable encoder: Linear($r$), [Linear($z$), Linear($z$)] (mean \& variance of $\mathbf{z}$);
    \item decoder: Linear($h$), ReLU, \{R-MP($h$), ReLU\}$ \times T$, Linear($h$), ReLU, [Linear($C$), Linear($C$)] (mean \& variance of $\mathbf{\hat{y}}$).
\end{enumerate}

Across all experiments, the Adam optimiser is used to maximise the ELBO (i.e. minimise the sum of the negative log-likelihood and KL-divergence in equation 5).

\subsubsection{PPISP} The hyperparameters used are $h=64$, $r=64$, $z=256$. Models were trained for 1000 epochs with a learning rate of $4 \times 10^{-5}$. We sample context and target points in the $10\%$--$50\%$ range.

\subsection{NP baseline}

The architecture of the NP consists of:
\begin{enumerate}
    \item encoder: Linear($h$), ReLU, Linear($h$), ReLU, Linear($r$);
    \item global latent variable encoder: same as for the MPNP;
    \item decoder: Linear($h$), ReLU, Linear($h$), ReLU, Linear($h$), ReLU, [Linear($C$), Linear($C$)] (mean \& variance of $\mathbf{\hat{y}}$).
\end{enumerate}

The Adam optimiser is also used here to maximise the ELBO.

\subsubsection{TUD} On Enzymes, the NP and NP-c hyperparameters are $h = 64, r = 128, z = 512$. On Proteins, we used $h = 64, r = 64, z = 512$ for the NP and $h = 64, r = 96, z = 288$ for the NP-c. On DHFR, COX2 and BZR, both NP and NP-c have $h = 64, r = 64, z = 512$. We trained both models for 400 epochs with learning rate $4e^{-5}$ on all datasets except for Enzymes, where we used 700 epochs. For all datasets, we sample $10\%$--$25\%$ context and (additional) target points.

\subsubsection{ShapeNet} The same hyperparameters were used for all tasks: $h = 64, r = 64, z = 512$. The NP was trained for 400 epochs on fixed-class and 500 epochs on mixed-class tasks, with $5\%$--$25\%$ context and (additional) target points and a learning rate of $4e^{-5}$.

\subsubsection{Cora} In both the transductive and few-shot settings,  $h = 64$, $r = 64$ and $z = N\times64$ for $N$-classes, matching the MPNP. The model is trained for 500 epochs in the transductive setting and 400 in the few-shot setting. A learning rate of $7\times10^{-5}$ is used in both cases and we sample context and target points in the $10\%$--$50\%$ range, matching the MPNP.

\subsubsection{CA} Changes are made to the NP architecture for the CA tasks to match the changes made to the MPNP for this task, with MP layers replaced with linear layers with $2h$ units to match the parameter count of the MP. Otherwise the parameters match that of the MPNP: $h=64$, $r=64$, $z=128$. The models are trained for 200 epochs with a learning rate of $1\times10^{-4}$.

\subsubsection{PPISP}
The hyperparameters used are $h=64$, $r=64$, $z=256$. Models were trained for 1000 epochs with a learning rate of $6 \times 10^{-5}$. We sample context and target points in the 10\%-50\% range, matching the R-MPNP.

\subsection{GNN baseline}

This model consists of 3 GCN\footnote{\url{https://pytorch-geometric.readthedocs.io/en/latest/modules/nn.html\#torch_geometric.nn.conv.GCNConv}} layers with learnable skip-connections; the operation of a layer is:
\begin{equation}
    \mathbf{h}_{t+1} = \text{ReLU}\big(\mathbf{W}_{\text{skip}}\mathbf{h}_t + \text{GCN}(\mathbf{h}_t)\big).
\end{equation}

We use $h = 64$ across all tasks and train the model for 500 epochs, with the Adam optimiser minimising the cross-entropy loss and a learning rate of $1e^{-4}$. The context and target ranges are as previously described, for each dataset. Note that this model does not make use of the context labels.

\subsection{R-GCN baseline}
The model consists of 3 RGCN\footnote{\url{https://pytorch-geometric.readthedocs.io/en/latest/modules/nn.html#torch_geometric.nn.conv.RGCNConv}} layers; the operation of a layer is:
\begin{equation}
    \mathbf{h}_{t+1} = \text{ReLU}\big(\text{RGCN}(\mathbf{h}_{t})\big).
\end{equation}

We use $h=64$ and train the model for 400 epochs, with the Adam optimiser minimising the cross-entropy loss and a learning rate of $7 \times 10^{-5}$. This model leverages edge features in the message-passing steps but does not make use of context labels.

\section{Numerical Results and Uncertainty Plots}
\label{apdx:numerical_results}

\begin{figure*}[t]
\centering
\includegraphics[trim = 130 20 45 20, clip, width=0.9\textwidth]{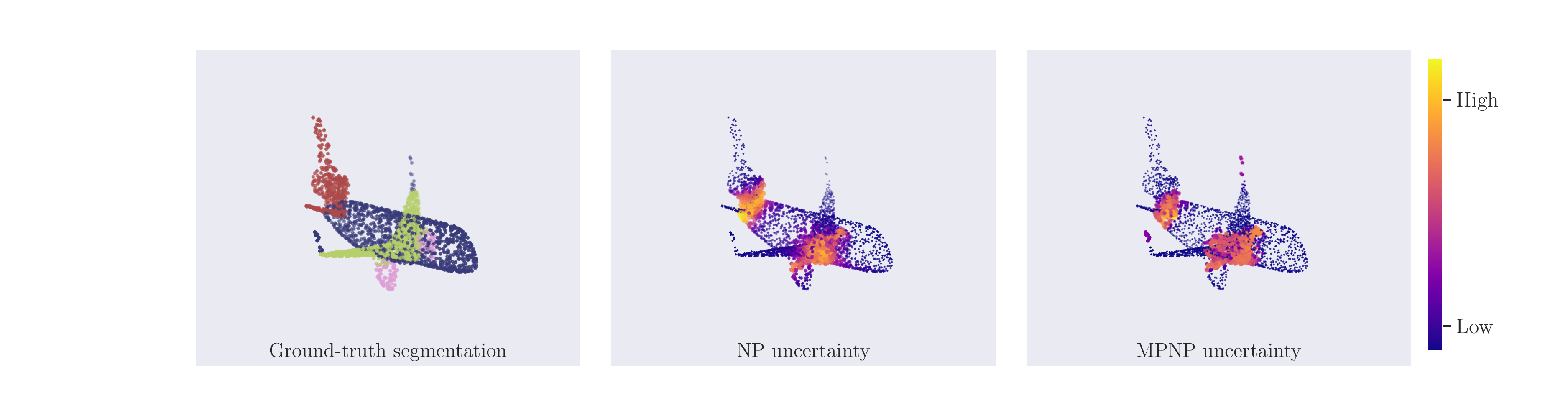}
\includegraphics[trim = 130 20 45 20, clip, width=0.9\textwidth]{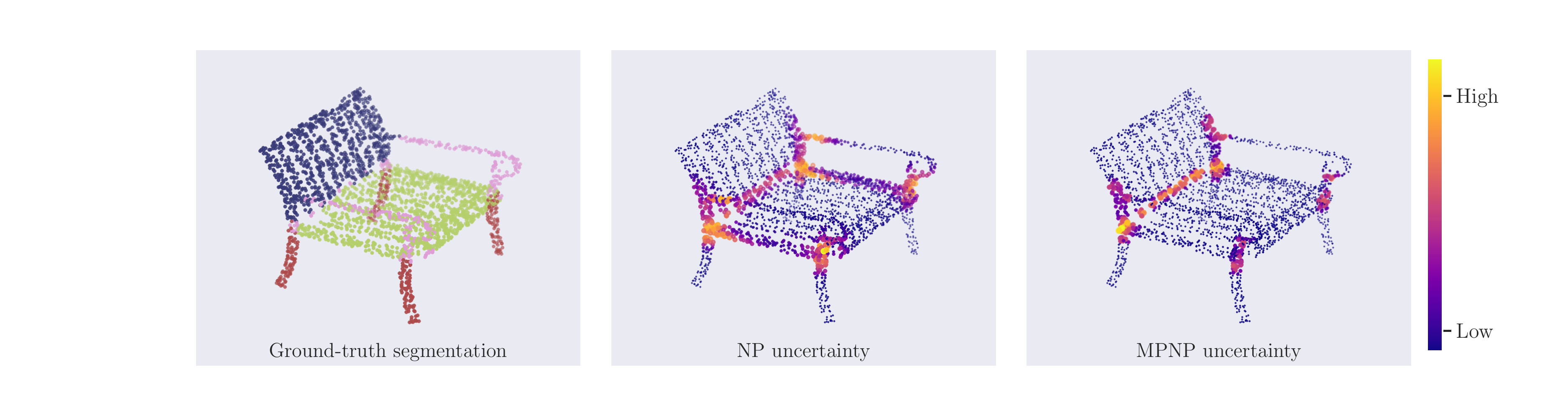}
\includegraphics[trim = 130 20 45 20, clip, width=0.9\textwidth]{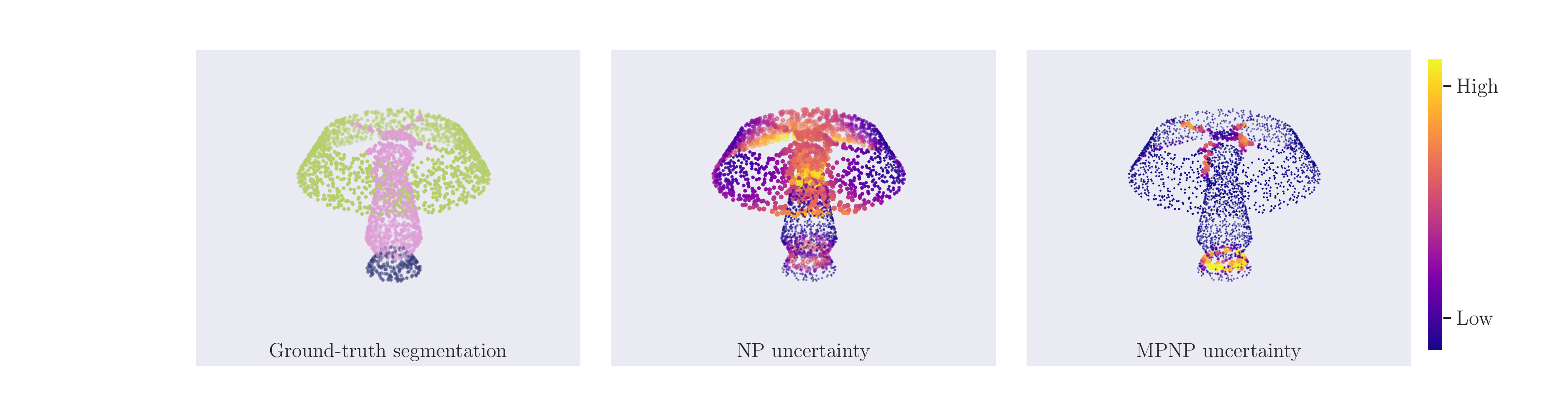}
\includegraphics[trim = 130 20 45 20, clip, width=0.9\textwidth]{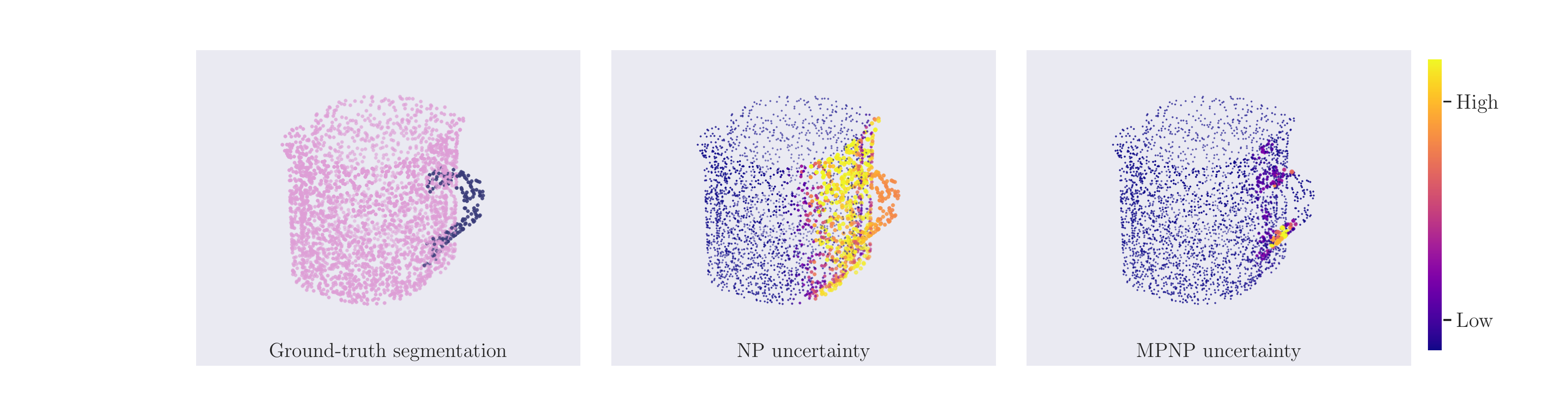}
\includegraphics[trim = 130 20 45 20, clip, width=0.9\textwidth]{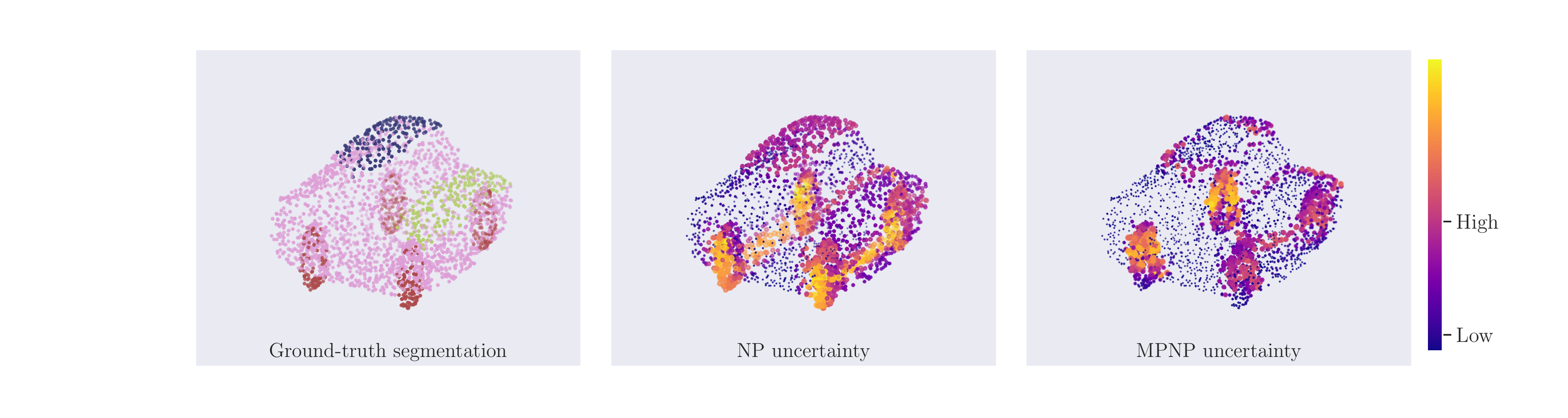}
\caption{Uncertainty visualisations on examples from the airplane, chair, lamp, mug and car categories. In each case the MPNP is able to better localise the uncertainty to semantically relevant locations (i.e. border regions). The NP tends to be uncertain in large simple volumes, having the entire handle side of the mug being very uncertain, for instance. Similar effects are seen for the top of the lamp, the car axles, and the edges of the chair seat. The airplane is generally harder, with borders between the wings, fuselage and engines occurring in a relatively compact region, though we still see better localisation in the tail.}
\label{fig:more-shapenet-viz}
\end{figure*}

In this section, we present the numerical results used to generate the CA and ShapeNet plots in the main text. Tables~\ref{tab:shapenet-single-mpnp-numresults}, ~\ref{tab:shapenet-single-np-numresults} and~\ref{tab:shapenet-single-gcn-labelprop-numresults} show the ShapeNet single-category performances, whereas Table~\ref{tab:CA-numresults} provides the Cellular Automata results. Figure~\ref{fig:more-shapenet-viz} illustrates additional uncertainty visualisations for other classes in ShapeNet, reinforcing the finding that the estimates produced by MPNPs are semantically relevant.

\begin{table*}[h]
    \centering
    \scriptsize
    \caption{Numerical mIoU results for the MPNP on ShapeNet single-category tasks ($\mu \pm \sigma$).}
    \begin{tabular}{c|ccccc}
    \toprule
         & 0.1\% & 1\% & 5\% & 10\% & 30\% \\
         \midrule
         Bag & $71.08 \pm 3.78$ & $75.12 \pm 1.44$ & $75.57 \pm 0.89$ & $76.05 \pm 0.13$ & $73.21 \pm 0.72$ \\
         Cap & $64.00 \pm 4.28$ & $68.76 \pm 4.32$ & $73.05 \pm 0.92$ & $69.42 \pm 1.34$ & $67.41 \pm 0.47$ \\
         Knife & $79.82 \pm 0.25$ & $87.34 \pm 1.47$ & $89.93 \pm 0.46$ & $90.39 \pm 0.34$ & $90.34 \pm 0.24$ \\
         Laptop & $90.39 \pm 0.31$ & $95.94 \pm 0.17$ & $96.71 \pm 0.19$ & $96.75 \pm 0.00$ & $97.07 \pm 0.12$ \\
         Mug & $75.90 \pm 1.37$ & $85.63 \pm 2.27$ & $87.80 \pm 1.02$ & $88.70 \pm 1.20$ & $88.14 \pm 0.02$ \\
         \midrule
         Earphone & $49.80 \pm 4.45$ & $57.14 \pm 1.99$ & $59.41 \pm 1.44$ & $55.59 \pm 1.22$ & $55.35 \pm 0.70$ \\
         Guitar & $77.95 \pm 0.61$ & $89.12 \pm 0.31$ & $92.33 \pm 0.25$ & $92.75 \pm 0.31$ & $93.17 \pm 0.11$ \\
         Pistol & $67.68 \pm 0.95$ & $82.51 \pm 0.60$ & $85.82 \pm 0.29$ & $85.57 \pm 0.46$ & $86.48 \pm 0.16$ \\
         Rocket & $54.61 \pm 0.21$ & $56.03 \pm 0.48$ & $60.78 \pm 0.74$ & $64.26 \pm 2.48$ & $62.90 \pm 0.04$ \\
         Skateboard & $41.67 \pm 2.01$ & $51.75 \pm 0.76$ & $55.10 \pm 0.13$ & $53.44 \pm 1.05$ & $52.33 \pm 0.15$ \\
         Table & $75.19 \pm 0.81$ & $83.64 \pm 0.30$ & $85.80 \pm 0.04$ & $85.94 \pm 0.01$ & $86.61 \pm 0.03$ \\
         \midrule
         Airplane & $58.32 \pm 0.82$ & $81.55 \pm 0.16$ & $86.68 \pm 0.06$ & $87.32 \pm 0.05$ & $87.90 \pm 0.00$ \\
         Car & $43.08 \pm 0.91$ & $69.02 \pm 1.45$ & $76.56 \pm 0.37$ & $78.02 \pm 0.12$ & $78.53 \pm 0.10$ \\
         Chair & $72.29 \pm 0.00$ & $86.73 \pm 0.32$ & $89.88 \pm 0.26$ & $90.22 \pm 0.00$ & $90.70 \pm 0.03$ \\
         Lamp & $61.80 \pm 1.31$ & $79.44 \pm 0.00$ & $84.03 \pm 0.15$ & $84.45 \pm 0.25$ & $84.89 \pm 0.01$ \\
         \midrule
         Motorbike & $27.54 \pm 1.36$ & $48.10 \pm 1.09$ & $53.94 \pm 0.16$ & $53.17 \pm 0.38$ & $53.76 \pm 0.02$ \\
    \bottomrule
    \end{tabular}
    \label{tab:shapenet-single-mpnp-numresults}
\end{table*}

\begin{table*}[h]
    \centering
    \scriptsize
    \caption{Numerical mIoU results for the NP on ShapeNet single-category tasks ($\mu \pm \sigma$).}
    \begin{tabular}{c|ccccc}
    \toprule
         & 0.1\% & 1\% & 5\% & 10\% & 30\% \\
         \midrule
         Bag & $52.62 \pm 0.00$ & $54.20 \pm 2.23$ & $52.87 \pm 0.35$ & $53.06 \pm 0.12$ & $53.46 \pm 0.13$ \\
         Cap & $45.08 \pm 6.01$ & $56.88 \pm 0.60$ & $55.21 \pm 0.40$ & $59.67 \pm 1.18$ & $57.94 \pm 0.71$ \\
         Knife & $72.84 \pm 0.33$ & $87.02 \pm 0.65$ & $89.49 \pm 0.12$ & $89.68 \pm 0.31$ & $89.12 \pm 0.20$ \\
         Laptop & $82.42 \pm 2.05$ & $93.49 \pm 0.24$ & $96.07 \pm 0.38$ & $96.46 \pm 0.12$ & $96.72 \pm 0.02$ \\
         Mug & $68.52 \pm 1.71$ & $80.95 \pm 0.41$ & $84.92 \pm 0.61$ & $84.58 \pm 0.97$ & $85.57 \pm 0.01$ \\
         \midrule
         Earphone & $36.42 \pm 7.28$ & $47.98 \pm 1.06$ & $47.94 \pm 0.68$ & $47.79 \pm 0.04$ & $49.04 \pm 0.24$ \\
         Guitar & $69.00 \pm 0.36$ & $83.41 \pm 0.64$ & $87.83 \pm 0.50$ & $88.12 \pm 0.16$ & $89.11 \pm 0.01$ \\
         Pistol & $63.84 \pm 2.02$ & $70.42 \pm 0.79$ & $70.64 \pm 0.15$ & $71.94 \pm 0.22$ & $71.79 \pm 0.17$ \\
         Rocket & $56.71 \pm 2.41$ & $59.76 \pm 0.69$ & $63.69 \pm 0.54$ & $62.50 \pm 0.58$ & $63.85 \pm 0.31$ \\
         Skateboard & $32.13 \pm 1.71$ & $41.84 \pm 0.07$ & $40.78 \pm 0.03$ & $40.36 \pm 0.19$ & $40.25 \pm 0.09$ \\
         Table & $76.53 \pm 0.21$ & $83.11 \pm 0.08$ & $84.18 \pm 0.08$ & $84.20 \pm 0.12$ & $84.26 \pm 0.01$ \\
         \midrule
         Airplane & $44.92 \pm 0.06$ & $78.05 \pm 0.22$ & $83.05 \pm 0.02$ & $83.65 \pm 0.02$ & $84.04 \pm 0.05$ \\
         Car & $38.86 \pm 0.43$ & $57.90 \pm 1.33$ & $63.89 \pm 0.08$ & $64.73 \pm 0.14$ & $65.55 \pm 0.46$ \\
         Chair & $69.68 \pm 1.14$ & $84.78 \pm 0.51$ & $87.35 \pm 0.10$ & $87.69 \pm 0.11$ & $87.81 \pm 0.01$ \\
         Lamp & $57.04 \pm 1.73$ & $71.88 \pm 0.54$ & $75.40 \pm 0.45$ & $75.49 \pm 0.19$ & $76.19 \pm 0.01$ \\
         \midrule
         Motorbike & $21.28 \pm 1.41$ & $25.44 \pm 0.05$ & $25.66 \pm 0.04$ & $25.69 \pm 0.16$ & $25.73 \pm 0.05$ \\
    \bottomrule
    \end{tabular}
    \label{tab:shapenet-single-np-numresults}
\end{table*}

\begin{table*}[h]
    \centering
    \scriptsize
    \caption{Numerical mIoU results for GCN and \texttt{labelprop} on ShapeNet single-category tasks ($\mu \pm \sigma$). Note that the GCN does not use the context labels and thus produces deterministic outputs.}
    \begin{tabular}{c|ccccc|ccccc}
    \toprule
         & \multicolumn{5}{c|}{GCN} & \multicolumn{5}{c}{\texttt{labelprop}} \\
         & \multicolumn{5}{c|}{0.1\% / 1\% / 5\% / 10\% / 30\%} & 0.1\% & 1\% & 5\% & 10\% & 30\% \\
         \midrule
         Bag & \multicolumn{5}{c|}{$69.76$} & $54.62 \pm 1.88$ & $70.10 \pm 4.35$ & $86.16 \pm 1.05$ & $90.45 \pm 1.10$ & $95.67 \pm 0.54$ \\
         Cap & \multicolumn{5}{c|}{$65.85$} & $47.76 \pm 5.07$ & $74.19 \pm 3.53$ & $84.97 \pm 0.49$ & $88.83 \pm 0.62$ & $93.43 \pm 0.41$ \\
         Knife & \multicolumn{5}{c|}{$79.61$} & $57.48 \pm 3.40$ & $88.82 \pm 0.56$ & $93.70 \pm 0.36$ & $95.01 \pm 0.24$ & $97.03 \pm 0.10$ \\
         Laptop & \multicolumn{5}{c|}{$94.23$} & $58.61 \pm 2.64$ & $88.16 \pm 0.64$ & $93.76 \pm 0.17$ & $95.46 \pm 0.12$ & $97.34 \pm 0.05$ \\
         Mug & \multicolumn{5}{c|}{$85.40$} & $47.44 \pm 1.48$ & $74.93 \pm 2.42$ & $88.15 \pm 0.45$ & $91.09 \pm 0.71$ & $94.19 \pm 0.15$ \\
         \midrule
         Earphone & \multicolumn{5}{c|}{$49.76$} & $36.35 \pm 2.29$ & $66.68 \pm 1.96$ & $78.45 \pm 1.21$ & $82.50 \pm 0.73$ & $88.24 \pm 0.27$ \\
         Guitar & \multicolumn{5}{c|}{$89.01$} & $37.85 \pm 1.69$ & $78.79 \pm 1.97$ & $92.06 \pm 0.20$ & $94.10 \pm 0.20$ & $96.44 \pm 0.09$ \\
         Pistol & \multicolumn{5}{c|}{$76.72$} & $39.80 \pm 1.22$ & $69.09 \pm 1.56$ & $83.25 \pm 0.94$ & $87.02 \pm 0.23$ & $92.39 \pm 0.21$ \\
         Rocket & \multicolumn{5}{c|}{$56.31$} & $38.43 \pm 3.14$ & $59.84 \pm 7.51$ & $80.95 \pm 0.52$ & $85.67 \pm 1.15$ & $91.53 \pm 0.37$ \\
         Skateboard & \multicolumn{5}{c|}{$57.19$} & $32.83 \pm 1.57$ & $56.59 \pm 1.35$ & $80.33 \pm 0.71$ & $84.91 \pm 0.78$ & $90.76 \pm 0.35$ \\
         Table & \multicolumn{5}{c|}{$76.54$} & $42.22 \pm 0.53$ & $68.88 \pm 0.27$ & $83.32 \pm 0.08$ & $86.83 \pm 0.06$ & $91.16 \pm 0.07$ \\
         \midrule
         Airplane & \multicolumn{5}{c|}{$79.50$} & $20.81 \pm 0.21$ & $60.48 \pm 0.33$ & $79.77 \pm 0.13$ & $84.50 \pm 0.07$ & $90.47 \pm 0.04$ \\
         Car & \multicolumn{5}{c|}{$71.95$} & $19.20 \pm 0.38$ & $45.91 \pm 0.99$ & $67.85 \pm 0.38$ & $75.33 \pm 0.31$ & $85.36 \pm 0.05$ \\
         Chair & \multicolumn{5}{c|}{$77.84$} & $33.06 \pm 0.62$ & $70.97 \pm 0.27$ & $87.04 \pm 0.12$ & $90.65 \pm 0.07$ & $94.65 \pm 0.02$ \\
         Lamp & \multicolumn{5}{c|}{$53.78$} & $58.61 \pm 2.64$ & $88.16 \pm 0.64$ & $93.76 \pm 0.17$ & $95.46 \pm 0.12$ & $97.34 \pm 0.05$ \\
         \midrule
         Motorbike & \multicolumn{5}{c|}{$46.18$} & $15.66 \pm 1.14$ & $38.46 \pm 1.28$ & $63.11 \pm 1.46$ & $74.05 \pm 0.51$ & $85.51 \pm 0.36$ \\
    \bottomrule
    \end{tabular}
    \label{tab:shapenet-single-gcn-labelprop-numresults}
\end{table*}

\begin{table*}[h]
    \centering
    \scriptsize
    \caption{Numerical accuracy results for the Cellular Automata tasks ($\mu \pm \sigma$).}
    \begin{tabular}{c|ccc|ccc}
    \toprule
         & \multicolumn{3}{c|}{MPNP} & \multicolumn{3}{c}{NP} \\
         & 10\% & 30\% & 100\% & 10\% & 30\% & 100\% \\
         \midrule
         Small-world & $\bf 88.14 \pm 0.82$ & $\bf 95.09 \pm 0.45$ & $\bf 97.33 \pm 0.57$ & $77.28 \pm 4.85$ & $78.97 \pm 4.87$ & $79.59 \pm 4.83$ \\
         Scale-free & $\bf 84.73 \pm 2.87$ & $\bf 93.18 \pm 3.25$ & $\bf 95.49 \pm 3.47$ & $74.84 \pm 0.73$ & $76.91 \pm 0.43$ & $77.57 \pm 0.47$ \\
         Voronoi & $\bf 83.13 \pm 2.77$ & $\bf 90.01 \pm 5.16$ & $\bf 92.39 \pm 6.26$ & $73.96 \pm 4.09$ & $76.09 \pm 4.46$ & $76.70 \pm 4.34$ \\
         Spherical Voronoi & $\bf 82.91 \pm 2.66$ & $\bf 91.68 \pm 3.90$ & $\bf 94.93 \pm 4.57$ & $74.00 \pm 4.12$ & $75.81 \pm 4.21$ & $76.53 \pm 3.86$ \\
         \midrule
         Life-like & $63.81 \pm	2.03$ & $65.40 \pm 2.73$ & $65.77 \pm 2.85$ & $61.38 \pm 0.77$ & $62.40 \pm 0.06$ & $62.62 \pm 0.03$ \\
         \toprule
         & \multicolumn{3}{c|}{Population mode} & \multicolumn{3}{c}{State mode} \\
         & 10\% & 30\% & 100\% & 10\% & 30\% & 100\% \\
         \midrule
         Small-world & $68.40 \pm 0.11$ & $69.53 \pm 0.13$ & $70.10 \pm 0.00$ & $80.34 \pm 0.13$ & $81.62 \pm 0.11$ & $82.11 \pm 0.00$ \\
         Scale-free & $66.54 \pm 0.42$ & $67.64 \pm 0.12$ & $68.25 \pm 0.00$ & $74.46 \pm 0.17$ & $76.30 \pm 0.12$ & $76.90 \pm 0.00$ \\
         Voronoi & $67.36 \pm 0.32$ & $68.71 \pm 0.04$ & $69.37 \pm 0.00$ & $76.54 \pm 0.17$ & $78.17 \pm 0.07$ & $78.74 \pm 0.00$ \\
         Spherical Voronoi & $66.09 \pm 0.29$ & $67.31 \pm 0.07$ & $67.91 \pm 0.00$ & $76.46 \pm 0.20$ & $78.13 \pm 0.04$ & $78.72 \pm 0.00$ \\
         \midrule
         Life-like & $62.08 \pm 0.07$ & $62.55 \pm 0.02$ & $62.69 \pm 0.00$ & $\bf 84.22 \pm 0.04$ & $\bf 84.48 \pm 0.02$ & $\bf 84.57 \pm 0.00$ \\
    \bottomrule
    \end{tabular}
    \label{tab:CA-numresults}
\end{table*}

\section{Inductive GNNs with Arbitrary Labelling}
When introducing the baselines in the Experiments section, we noted that the expected performance of inductive GNNs in the arbitrary labelling setting is no better than chance. This is because the predictions of such a model do not depend on the labelling scheme and for any particular labelling of a task we can produce a set of equivalent tasks by permuting the labels. First consider the two class case: outputs are either 1 or 2 and labels are either $\mathrm{A}$ or $\mathrm{B}$, giving the mutually exclusive, collectively exhaustive groups $\{1\mathrm{A}\},\{1\mathrm{B}\},\{2\mathrm{A}\},\{2\mathrm{B}\}$, which we normalise to sum to 1 by dividing by the number of examples. In the case that $(\mathrm{A}, \mathrm{B}) = (1,2)$, the accuracy is:

\begin{equation*}
    \mathrm{acc}_{\mathrm{A}\mathrm{B}} = \{1\mathrm{A}\}+\{2\mathrm{B}\}
\end{equation*}
and if the labels are permuted:
\begin{equation*}
    \mathrm{acc}_{\mathrm{B}\mathrm{A}} = \{2\mathrm{A}\}+\{1\mathrm{B}\}
\end{equation*}

which average to:

\begin{equation*}
    \mathrm{acc}_{\mathrm{mean}} = \frac{\{1\mathrm{A}\}+\{2\mathrm{B}\} + \{2\mathrm{A}\}+\{1\mathrm{B}\}}{2} = \frac{1}{2}.
\end{equation*}
Generalising, outputs are in ${1,...,\mathrm{N}}$ and labels in $\{\mathrm{A},...,\Omega\}$, for the matrix of pairs:
\begin{equation*}
\begin{bmatrix} 
    1\mathrm{A} & \dots & 1\Omega \\
    \vdots & \ddots & \\
    \mathrm{N}\mathrm{A} &        & \mathrm{N}\Omega
    \end{bmatrix}
\end{equation*}

with the sum of all these elements being 1. There are $\mathrm{N}!$ permutations of the arbitrary labelling, and therefore $\mathrm{N}!$ equivalent tasks. Each term in the matrix appears in $(\mathrm{N}-1)!$ accuracy sums (with that term fixed, there are $\mathrm{N}-1$ free terms with $(\mathrm{N}-1)!$ permutations), so the mean accuracy is:
\begin{equation*}
    \mathrm{acc}_{\mathrm{mean}} = \frac{(\mathrm{N}-1)!\left(1\mathrm{A}+\dots+\mathrm{N}\Omega\right)}{\mathrm{N}!}=\frac{(\mathrm{N}-1)!}{\mathrm{N}!} = \frac{1}{\mathrm{N}}.
\end{equation*}

\section{An MPNP Solution to the Life-like Family}
\label{apdx:lifelike_solution}
The Life-like rules can be viewed as 18 separate rules that act in parallel, one for each neighbourhood count (9) for each state (2) and hence the $2^{18}$ variants noted in the main text (experimental section). A solution can be built using the concatenation encoder where the first steps describe the situation being observed at a given node as a one-hot encoding in an 18-element vector, and then summarises these using a max aggregator (or sum or mean with corrections later) and then concatenating by whether the cell lives or dies (i.e. concatenate by class). The max aggregation gives all the observed conditions that lead to a cell being alive in the next generation, and all those that lead to a cell being dead\footnote{This could be compressed further using the fact that the rules are deterministic and do not overlap.}. This representation is then used without modification as the latent variable. The decoder first extracts the observation to the format used by the encoder and then compares it with the latent variable, if it matches a condition found in the living-half of the latent variable, then the cell is alive in the next generation, if it matches a condition in the dead-half then the cell dies or stays dead.

The non-obvious parts are producing a one-hot encoding from a scalar (the neighbourhood count is produced simply by the MP) and checking the decoder observation against the latent variable. One-hot encodings of length $N$ can be produced using Maxout layers as follows. First using a 2-pool Maxout as:
\begin{equation*}
    \textup{Maxout}(x) = \textup{max}_j\big((W_1x+b_1)_j,(W_2x+b_2)_j\big),
\end{equation*}
setting $W_1 = -1$ and $W_2 = 1$, $b_1 = (-1,0,...,N-2)$ and $b_2=(-1,-2,...,-N)$. The elements of this function take the form of the max of $(-x+j-1)$ and $(x-j-1)$ which is a `v' with unit slopes centred at $j$ with a minimum value of $-1$. If we follow the Maxout with a linear layer ($-I$) and a ReLU activation, we can first flip the `v' and then flatten the edges to give a triangular hat centred at $j$ with height 1. Thus, if $x=2$ the first element (zeroth) is 0, the second element is 0, the third element is 1 and the rest are 0s. In this way, the first parts of the encoder and decoder can accurately represent the observed states. To compare an observation against the latent variable, we can take the sum of the observation and latent and subtract $1$s (i.e. an AND).
\end{appendices}

\end{document}